\documentclass[runningheads]{llncs}

\usepackage{eccv}

\usepackage{eccvabbrv}

\usepackage{graphicx}
\usepackage{booktabs}
\usepackage[accsupp]{axessibility}  
\usepackage{amsmath}
\usepackage{algorithm}
\usepackage{algpseudocode}
\usepackage{wrapfig}
\usepackage{multirow}
\usepackage{booktabs}       
\usepackage{multirow}       
\usepackage{array}          
\usepackage{graphicx}       
\usepackage[utf8]{inputenc}
\usepackage[table]{xcolor}  
\usepackage[most]{tcolorbox} 
\usepackage{enumitem}       

\usepackage{arydshln}       
\usepackage{amsmath}        
\usepackage{amssymb}        
\usepackage{graphicx}
\usepackage{subcaption}
\usepackage{pifont}
\newcommand{\xmark}{\textcolor{red!70!black}{\ding{55}}}   
\newcommand{\cmark}{\textcolor{green!60!black}{\ding{51}}} 

\definecolor{graybg}{HTML}{F2F2F2}  
\definecolor{gainbg}{HTML}{E2F0D9}  
\definecolor{generalbg}{RGB}{235, 245, 255}  
\definecolor{deltagreen}{HTML}{008000}

\definecolor{deltagreen}{HTML}{008000} 
\definecolor{graybg}{HTML}{F2F2F2}
\definecolor{gainbg}{HTML}{E7F3E7}
\usepackage{arydshln}

\usepackage[pagebackref,breaklinks,colorlinks,citecolor=eccvblue]{hyperref}
\usepackage[capitalize]{cleveref} 
\usepackage{orcidlink}

\begin{document}

\title{How Far Are Video Models from True Multimodal Reasoning?} 


\author{
Xiaotian Zhang\inst{1,2} \textsuperscript{*,\S} \and
Jianhui Wei\inst{1,2} \textsuperscript{*,\S} \and
Yuan Wang\inst{1} \textsuperscript{*} \and
Jie Tan\inst{1} \and \\
Yichen Li\inst{1} \and
Yan Zhang\inst{2} \textsuperscript{\ddag} \and
Ziyi Chen\inst{2} \and
Daoan Zhang\inst{2} \and
Dezhi Yu\inst{2} \and \\
Wei Xu\inst{2} \and
Songtao Jiang\inst{1} \and
Zuozhu Liu\inst{1} \textsuperscript{\dag}
}

\authorrunning{X. Zhang et al.}

\institute{Zhejiang University \and ByteDance \\
\textsuperscript{*}Equal contribution. 
\textsuperscript{\dag}Corresponding author.
\textsuperscript{\ddag}Project leader. \\
\textsuperscript{\S}Work done during internship at ByteDance.
}

\maketitle

\begin{abstract}

Despite remarkable progress toward general-purpose video models, a critical question remains unanswered: how far are these models from achieving true multimodal reasoning? Existing benchmarks fail to address this question rigorously, as they remain constrained by straightforward task designs and fragmented evaluation metrics that neglect complex multimodal reasoning. To bridge this gap, we introduce CLVG-Bench, an evaluation framework designed to probe video models' zero-shot reasoning capabilities via \textbf{C}ontext \textbf{L}earning in \textbf{V}ideo \textbf{G}eneration. CLVG-Bench comprises more than 1,000 high-quality, manually annotated metadata across 6 categories and 47 subcategories, covering complex scenarios including physical simulation, logical reasoning, and interactive contexts. To enable rigorous and scalable assessment, we further propose an \textbf{A}daptive \textbf{V}ideo \textbf{E}valuator (AVE) that aligns with human expert perception using minimal annotations, delivering interpretable textual feedback across diverse video context tasks. Extensive experiments reveal a striking answer to our central question: while state-of-the-art (SOTA) video models, such as Seedance 2.0,  demonstrate competence on certain understanding and reasoning subtasks, they fall substantially short with logically grounded and interactive generation tasks (achieving success rates $<25\%$ and $\sim 0\%$, respectively), exposing multimodal reasoning and physical grounding as critical bottlenecks. By systematically quantifying these limitations, the proposed method provides actionable feedbacks and a clear roadmap toward truly robust, general-purpose video models. CLVG-Bench and code are released \href{https://github.com/Monncyann/CLVG-Bench}{here}.
\end{abstract}

\section{Introduction}
\label{sec:intro}

Driven by large-scale training on web-scale data with generative objectives~\cite{brown2020language,wei2022emergent,jiang2025hulu}, video models have demonstrated groundbreaking zero-shot capabilities, evolving from mere instruction following to complex understanding and reasoning~\cite{video_reasoner,seedance1_5,kling_omni,wang2025v2t}. Specifically, traditional reference-based video generation has transitioned from simple, single-reference conditioning~\cite{fulldit,moviegene,wan,hunyuanvideo,anyv2v,unic,vace} to accommodating multi-subject and multimodal references~\cite{hu2024instruct,univideo}. Concurrently, the primary focus of these tasks is shifting from prioritizing sheer visual fidelity to emphasizing broader reasoning capabilities. Despite this immense potential and the surging interest in video reasoning, the community still lacks a systematic formulation and comprehensive benchmarking for these emerging paradigms. To address this, we formalize the task of reasoning-driven video generation, conditioned on arbitrary combinations of modalities (text, image, audio, and video), as \textbf{Context Learning in Video Generation}.

\begin{figure*}[t]
\centering
  \includegraphics[width=1\textwidth]{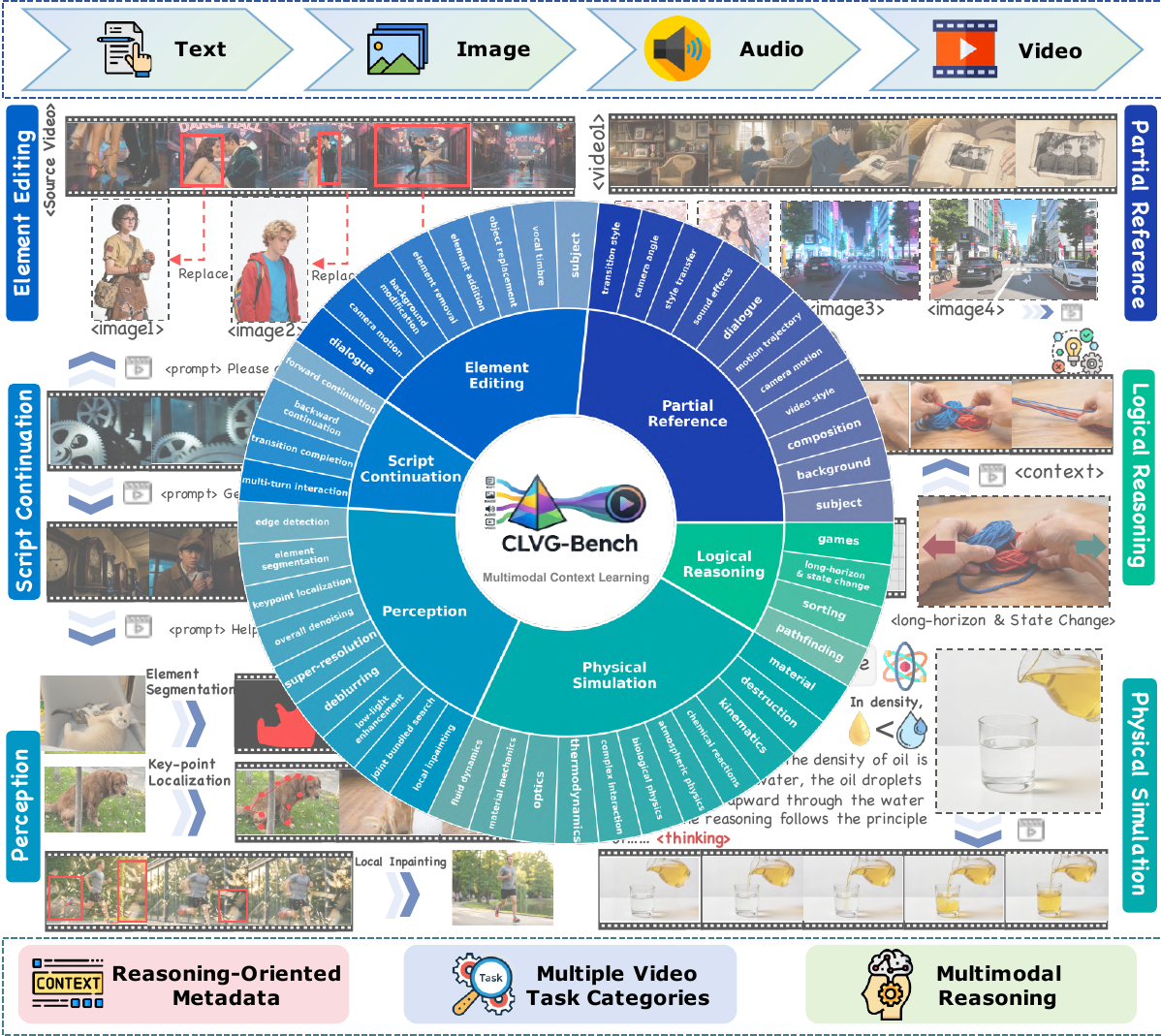}
  \caption{\textbf{Overview of CLVG-Bench}. Our framework evaluates video models via Context Learning in Video Generation across arbitrary modalities (Text, Image, Audio, Video). It systematically probes reasoning capabilities across \textbf{6} categories and \textbf{47} subcategories, which provides a roadmap for general-purpose video models.}
  \label{fig:CLVG-Bench}

\end{figure*}

Second, video quality evaluation has transitioned from fragmented scoring metrics~\cite{vbench} toward fine-grained and unified textual interpretations. 
Previous evaluation paradigms typically rely on task-specific metric combinations~\cite{veditbench,flownet,zhang2018unreasonable,huang2023vbench,liu2023evalcrafter,liu2023fetvbenchmarkfinegrainedevaluation,ji2024t2vbench,sun2025t2v} or reward models trained on multi-dimensional preference datasets~\cite{he2024videoscore,xu2024visionrewardfinegrainedmultidimensionalhuman,he2025videoscore2thinkscoregenerative,bansal2025videophy2challengingactioncentricphysical,bansal2024videophyevaluatingphysicalcommonsense}. 
However, fragmented and coarse metrics often fail to provide actionable textual feedback, while reward model training is frequently bottlenecked by the scarcity of high-quality data and remains susceptible to reward hacking~\cite{eisenstein2024helpingherdingrewardmodel}. 
To circumvent these issues, recent agent-based frameworks~\cite{yang2025videogenevalagentbasedvideogeneration,han2025videobenchhumanalignedvideogeneration,wei2026univbenchunifiedevaluationvideo} have been adapted for unified video evaluation, offering both quantitative scores and qualitative feedback. 
Nonetheless, their reliance on hand-crafted system prompts and static in-context examples severely constrains their generalization and reproducibility. They primarily focus on visual fidelity while overlooking intrinsic logical reasoning within videos. Consequently, efficiently aligning these agents to new video evaluation tasks with minimal expert supervision remains a critical yet underexplored problem.

We introduce \textbf{CLVG-Bench}, designed to comprehensively evaluate video models' ability to perform zero-shot reasoning on multimodal contexts. We systematically define 6 fundamental categories and 47 subcategories, and have carefully curated more than 1,000 high-quality metadata entries with video reasoning-oriented contexts, including text instruction, multi-shot and multi-subject videos, reference images and audio. To prevent the copyright issues and data contamination in evaluation, all metadata (\eg videos, images) are manually curated. We further propose an \textbf{A}daptive \textbf{V}ideo \textbf{E}valuator (\textbf{AVE}), leveraging an Automatic Prompt Optimization (APO) framework equipped with a novel semantic matching function for open-ended text evaluation. This mechanism empowers the evaluation agent to adapt to specific tasks using minimal expert annotations while simultaneously generating detailed textual feedback. Consequently, AVE serves as a interpretable metric for assessing the reasoning capabilities of video models, supporting scalable evaluation across diverse models and tasks, and facilitating fine-grained error attribution.

Experimental results reveal significant performance variations among current video models across different reasoning tasks. In particular, these models exhibit considerable shortcomings in tasks related to physical reasoning, logical inference, and interactive generation. This highlights a critical gap in their capability to simulate and reason about real-world dynamics.

Further analysis reveals that explicit context understanding provided by vision-language models aids video models in generating physically plausible content. This indicates that current video models lack the intrinsic ability to autonomously infer physical causality within instructions. Moving forward, future architectures should integrate comprehension and generation components more seamlessly to achieve robust, reliable reasoning capabilities in complex scenarios.

Our major contributions are threefold:

\begin{itemize}
    \item We introduce CLVG-Bench, a novel evaluation framework that abstracts video generation tasks into the definition of context learning video generation, and systematically evaluates the capability of current video models in simulating and reasoning about real-world dynamics.
    \item We propose the Adaptive Video Evaluator, a flexible evaluation framework designed for open-ended generation tasks. This evaluator dynamically adjusts based on a minimal set of human annotations, offering a versatile approach to assess tasks with varying contexts.
    \item Our study uncovers the limitations of current video models in multimodal reasoning. We advocate for a tighter integration of understanding and generation to enhance model performance in complex video generation scenarios.
\end{itemize}

\section{Related Works}

\subsection{Evolution of Video Models}

Video models have evolved from early diffusion frameworks \cite{1-wang2023modelscope,wu2024lamp,wan} to advanced Diffusion Transformer (DiT) architectures \cite{kling_omni,vidu2024,google2025veo3,kong2024hunyuanvideo,pixverse2023,zhao2024dynamic}, achieving superior cinematic quality, physical simulation, and efficiency \cite{hacohen2026ltx}. To bridge understanding and generation, unified models \cite{8-team2024chameleon,11-wang2024emu3} now integrate cross-modal decoding within single architectures. Recent works advance this bidirectional paradigm through joint audio-visual DiTs \cite{seed2} and integrating LLMs with 3D visual tokenizers \cite{13-chen2025janus,15-liao2025mogao,univideo}, driving holistic multimodal intelligence.

\subsection{Video Benchmark}

Video benchmarks have shifted from discriminative understanding to generative reasoning. Early efforts \cite{xu2016msr,caba2015activitynet,johnson2017clevr} focused on temporal and diagnostic reasoning, a trajectory recently extended by VBVR \cite{wang2026very} to probe complex logical chains. However, these web-scraped datasets face copyright and contamination risks while prioritizing perception over synthesis. Consequently, generation and editing benchmarks \cite{huang2024vbench,ren2026videoworld,sun2025ve,he2025openve} established aesthetic and instruction-following frameworks. Although recent works \cite{12-ye2025unic,vace} support reference-guided editing, they are often limited to single-shot scenarios and lack cross-shot consistency. More recently, the field has gravitated toward physical and spatial grounding. Benchmarks such as PhyGenBench \cite{meng2024towards}, RBench-V \cite{guo2025rbench}, and SpatialViz-Bench \cite{wangspatialviz} evaluate physical commonsense and 3D spatial relationships. Despite these advances, existing frameworks typically evaluate tasks in isolation, lacking a unified paradigm for multimodal context video generation.

\subsection{Video Evaluation Methods}
Existing video generation evaluation methodologies can be broadly categorized into three paradigms: heuristic foundation metrics, trained reward models, and agent-based LLM-as-judge frameworks. Heuristic metrics based on vision foundation models \cite{huang2023vbench,liu2023evalcrafter,ji2024t2vbench,sun2025t2v} are widely adopted for their computational efficiency. For example, CLIP-Score and its variants \cite{hessel2022clipscorereferencefreeevaluationmetric,liu2023fetvbenchmarkfinegrainedevaluation} for text-to-video consistency measurement. While efficient, these methods only yield coarse holistic scores and lack the fine-grained feedback required for detailed error analysis. The second paradigm focuses on trained specialized reward models. For example: \cite{he2024videoscore} which uses Video-LLMs to generate human-aligned multi-dimensional regression scores. However, these models require substantial expert-annotated data, making them hard to scale with rapidly evolving video tasks. Recently, agent-based LLM-as-judge frameworks have emerged as flexible evaluation solutions \cite{yang2025videogenevalagentbasedvideogeneration,han2025videobenchhumanalignedvideogeneration,wei2026univbenchunifiedevaluationvideo}. Despite their versatility, these agent-based approaches rely on expert-crafted system prompts and in-context examples, limiting reproducibility and scalability. To address these inherent limitations, our AVE leverages prompt optimization to iteratively refine the evaluation agent's system prompt, enabling dynamic adaptation to diverse video tasks with minimal expert annotations.



\section{Method}

\begin{table*}[t]
\centering
\small
\setlength{\tabcolsep}{5pt}
\renewcommand{\arraystretch}{1.0}
\caption{\textbf{Comparison of CLVG-Bench with existing video benchmarks.} Our benchmark support arbitrary combinations of multimodal inputs (Text, Image, Audio, and Video) while covering the broadest range of reasoning-oriented generation tasks, from element editing to multi-turn interaction.}

\resizebox{1.0\textwidth}{!}{
\begin{tabular}{lcccccc}
\toprule
\multicolumn{7}{l}{\textbf{Tasks} : \ding{172}: E.E. \ \ \ \ \  \ding{173}: P.R. \ \ \ \ \ \ding{174}: S.C. \ \ \ \ \ \ding{175}: P.S. \ \ \ \ \ \ding{176}: Perc. \ \ \ \ \ \ding{177}: L.R. \ \ \ \ \ \ding{178}: Interaction} \\
\hline
\hline
\textbf{Benchmark} & 
\textbf{Text} & 
\textbf{Image} & 
\textbf{Audio} &
\textbf{Video} &
\textbf{Reasoning-oriented} &
\textbf{Categories}\\
\midrule
VBench~\cite{huang2024vbench} & \cmark & \xmark & \xmark & \xmark & \xmark & \ding{173}  \\
VE-Bench~\cite{sun2025ve} & \cmark & \xmark & \xmark & \cmark & \xmark & \ding{172} \ding{173}  \\
UNIC~\cite{ye2025unic} & \cmark & \cmark & \xmark & \cmark & \xmark & \ding{172} \ding{173}  \\
VACE-Bench~\cite{jiang2025vace} & \cmark & \cmark & \xmark & \cmark & \xmark & \ding{172} \ding{173}  \\
FiVE-Bench~\cite{li2025five} & \cmark & \xmark & \xmark & \cmark & \xmark & \ding{172} \ding{173}  \\
UniVBench~\cite{wei2026univbenchunifiedevaluationvideo} & \cmark & \cmark & \xmark & \cmark & \xmark & \ding{172} \ding{173}  \\
RBench-V~\cite{guo2025rbench} & \cmark & \cmark & \xmark & \xmark & \cmark & \ding{177}  \\
Video-CraftBench~\cite{videoworld} & \cmark & \cmark & \xmark & \cmark & \cmark & \ding{177}  \\
OpenVE-3M~\cite{he2025openve} & \cmark & \xmark & \xmark & \cmark & \xmark & \ding{172} \ding{173}  \\
VBVR-Bench~\cite{vbvr} & \cmark & \cmark & \xmark & \cmark & \cmark & \ding{175} \ding{176} \ding{177}  \\
\midrule
\rowcolor{gray!15}
\textbf{CLVG-Bench} & \cmark & \cmark & \cmark & \cmark & \cmark & \ding{172}$\sim$\ding{178}\\
\bottomrule
\end{tabular}
}
\small \textbf{Note:} \textbf{E.E.}: Element Editing; \textbf{P.R.}: Partial Reference; \textbf{S.C.}: Script Continuation; \textbf{P.S.}: Physical Simulation; \textbf{Perc.}: Perception; \textbf{L.R.}: Logical Reasoning.

\label{tab:longbench_comparison}
\end{table*}

\subsection{Dataset Construction}

\noindent \textbf{Metadata Definition.} Irrespective of the specific downstream tasks, user inputs can be systematically categorized based on their modality formats. As delineated in~\cref{tab:longbench_comparison}, in a distinct departure from prior works, CLVG-Bench abstracts user prompts into versatile combinations of four foundational modalities: text, image, audio, and video. Furthermore, adhering to a reasoning-centric paradigm, we incorporate highly intricate reference-guided editing scenarios, including multi-shot and multi-subject tasks. Based on partial definitions from prior works~\cite{video_reasoner,univideo,videoworld}, we categorize context in video generation tasks into six fundamental categories comprising 47 fine-grained subcategories. The traditional elements editing, partial reference, and script continuation are seamlessly integrated with more context-aware physical simulation, perception, logical reasoning, and multi-turn interaction to formulate the comprehensive metadata architecture of CLVG-Bench (we temporarily incorporate the multi-turn interactive generation into the script continuation.~\cref{fig:CLVG-Bench} illustrates the detailed categories and provides several example demonstrations).
For examples of each specific subcategory, please refer to the supplementary material.

\noindent \textbf{Metadata Synthesis.} Directly using copyrighted web data may not only contaminate the evaluation data but also introduce a series of potential privacy concerns. To address these issues, all content in CLVG-Bench is human-created and carefully curated, ensuring that it is entirely free of copyright restrictions. This design enables a fair evaluation of in-context video generation capabilities and instruction following performance without legal constraints.

\begin{wrapfigure}{htbp}{0.55\textwidth}
  \centering
  \includegraphics[width=0.54\textwidth]{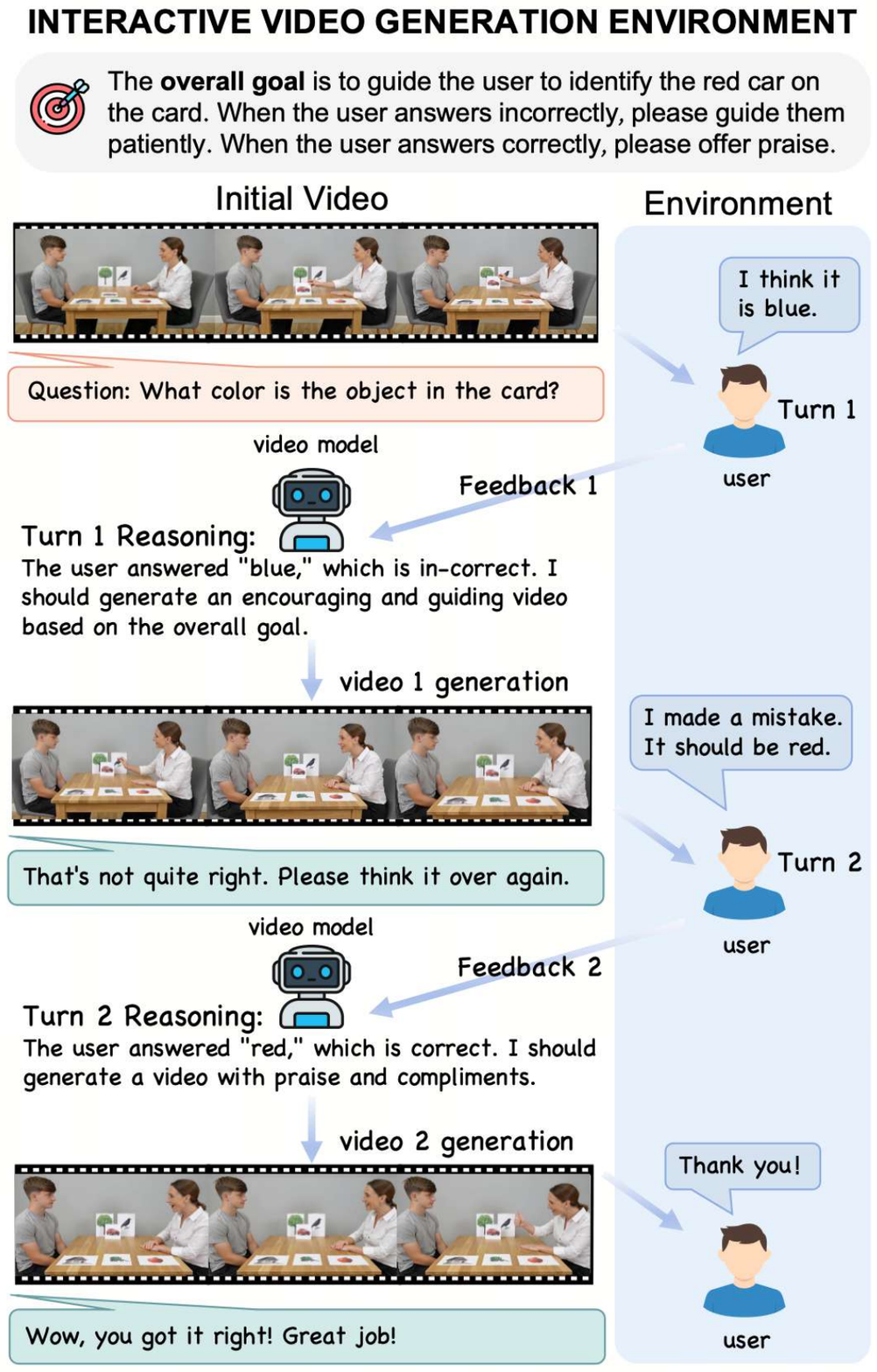}
  \caption{An environment for testing the multi-turn interactive generation capabilities of video models, where the model is expected to infer and predict a suitable video script based on the history context and user feedback.
}
  \label{fig:inactive_env}
\end{wrapfigure}

As described above, the metadata consists of four components: text (also called context), images, video, and audio. The context serves as the fundamental basis for reference to video generation. We use Seed 2.0 Pro~\cite{seed2} to generate reference contexts covering all subcategories under the following principles:
(1) extract themes for all six categories and their subcategories;
(2) expand the original theme and further enrich it into realistic user-oriented requirement descriptions.
Subsequently, three annotators independently verify the generated reference contexts according to their assigned categories, examining semantic completeness and requirement accuracy. Gemini 3 Pro~\cite{gemini3pro_modelcard_2025} provides additional automated verification by cross-checking factual statements. Any context that fails to adhere to the predefined category specifications undergoes collaborative review, where annotators discuss discrepancies and produce a corrected version.

Before generating the metadata videos, we follow three steps to produce content-rich and diverse video scripts:
(1) Based on Andrew \etal~\cite{andrew1984concepts} and Bordwell \etal~\cite{bordwell2008film}, we define five fundamental elements for video generation: subject, vocal characteristics of the subject, number of shots, camera movement, and video genre (For the data distribution, please refer to the supplementary materials).
(2) For each video script, we randomly assign five labels (sampled without replacement to ensure a uniform distribution) to construct its core elements.
(3) We then invoke Seed 2.0 Pro to generate a complete and coherent script for each case based on the assigned labels, ensuring that it is directly usable for video generation.
Finally, we use Seedance 2.0 to generate the video conditioned on the full script.

UniVBench \cite{wei2026univbenchunifiedevaluationvideo} contains a diverse set of reference images that are free from copyright concerns. We manually curated and filtered out samples that might cause style drift or introduce ambiguous references. The remaining images were retained and used as metadata images. 
For the metadata audio, we curate and collect audio data from the Seed-VC open-source repository~\cite{seed-vc}, filtering out segments with noisy backgrounds and low quality.

\noindent \textbf{Multi-turn Interaction.} While some prior work has explored the physical simulation and logical reasoning capabilities of video models~\cite{handcraft,vbvr,videoworld,wu2026visual}, few studies focus on the multi-turn interactive generation of videos. This task requires video models to not only understand but also iteratively predict video scripts based on user feedback, rather than simply following user input instructions. Such capabilities are present in current LLMs and VLMs, but are still lacking in video models. In CLVG-Bench, we introduce the novel task of multi-turn interactive video generation and design a series of interactive feedback environments. As illustrated in~\cref{fig:inactive_env}, inspired by early education scenarios, we require video models to assume the role of a teacher in the overall goal. Given an initial video, the environment provides random feedback based on its content. The baseline model is expected to reason and generate the correct video script in response to the feedback turn by turn.

\begin{figure*}[htbp]
\centering
  \includegraphics[width=1\textwidth]{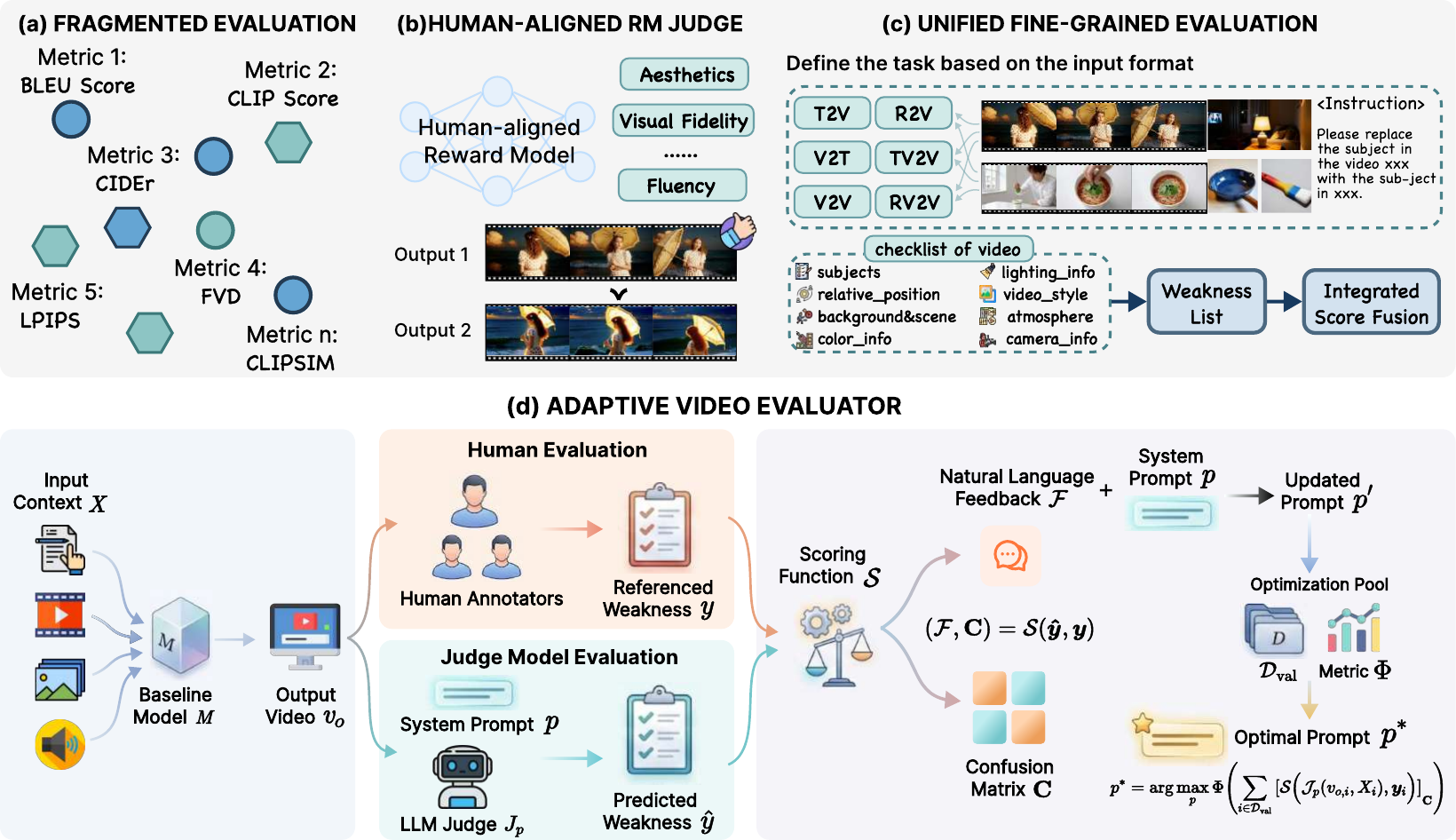}
  \caption{The top tier represents the previous evaluation methodologies: (a) fragmented evaluation, (b) human-aligned reward models, and (c) unified fine-grained evaluation. The lower tier illustrates the pipeline of the Adaptive Video Evaluator.}
  \label{fig:AVE_pipeline}
\end{figure*}

\subsection{Evaluation Pipeline}





\noindent \textbf{Problem Formulation.} Let the input context be defined as a tuple $X \in \{\text{text}, v_r, \text{image}, \text{audio}\}$, where $v_r$ represents a reference video. A baseline model $\mathcal{M}$ generates an output video $v_o$ based on the input, denoted as $v_o = \mathcal{M}(X)$. 

Human annotators assess $v_o$ based on context $X$ and provide a referenced set of noticeable weaknesses, denoted as $\boldsymbol{y}$. Concurrently, a VLM-based judge agent $\mathcal{J}_p$, conditioned on a system prompt $p$, predicts a corresponding set of weaknesses $\boldsymbol{\hat{y}} = \mathcal{J}_p(v_o, X)$. To evaluate the judge agent, a semantic matching function $\mathcal{S}$ compares the prediction $\boldsymbol{\hat{y}}$ against the referenced $\boldsymbol{y}$, yielding natural language feedback $\mathcal{F}$ and a normalized instance-level confusion matrix $\mathbf{C} \in [0, 1]^{1 \times 4}$ (comprising TP, TN, \etc ):
\begin{align}
    (\mathcal{F}, \mathbf{C}) = \mathcal{S}(\boldsymbol{\hat{y}}, \boldsymbol{y})
    \label{eq:evaluation}
\end{align}
The feedback $\mathcal{F}$ details the discrepancies between $\boldsymbol{\hat{y}}$ and $\boldsymbol{y}$, while the elements of $\mathbf{C}$ sum to 1. The matching logic is shown in \cref{alg:scoring_function}.

To optimize the judge agent $\mathcal{J}_p$, we collect a batch of annotated dataset $\mathcal{D} = \{(X_i, v_{o,i}, \boldsymbol{y}_i)\}_{i=1}^N$. This dataset is further partitioned into training, validation, and testing subsets, denoted as $\mathcal{D}_{\text{train}}$, $\mathcal{D}_{\text{val}}$, and $\mathcal{D}_{\text{test}}$, respectively. 

During the optimization phase, we utilize $\mathcal{D}_{\text{train}}$ and $\mathcal{D}_{\text{val}}$ to explore and verify prompt candidates. By aggregating the instance-level confusion matrices across the validation set into a global confusion matrix, our objective is to find the optimal prompt $p^*$ that maximizes a specific evaluation metric $\Phi(\cdot)$ (e.g., F1 score or Recall minus FPR):

\begin{equation}
    p^* = \arg\max_{p} \Phi \left( \sum_{i \in \mathcal{D}_{\text{val}}} \left[ \mathcal{S} \Big( \mathcal{J}_p(v_{o,i}, X_i), \boldsymbol{y}_i \Big) \right]_{\mathbf{C}} \right)
    \label{eq:objective_compact}
\end{equation}

Finally, the optimal prompt $p^*$ is frozen and evaluated on the unseen test set $\mathcal{D}_{\text{test}}$ to measure the generalized performance of the judge agent:

\begin{align}
    \text{Score}_{\text{test}} &= \Phi \left( \sum_{i \in \mathcal{D}_{\text{test}}} \left[ \mathcal{S} \Big( \mathcal{J}_{p^*}(v_{o,i}, X_i), \boldsymbol{y}_i \Big) \right]_{\mathbf{C}} \right) \label{eq:test_score}
\end{align}

\begin{algorithm}[t] 
\caption{Semantic Matching $\mathcal{S}$}
\label{alg:scoring_function}
\small 
\begin{algorithmic}[1] 
\Require referenced and predicted weaknesses $\boldsymbol{y}$, $\hat{\boldsymbol{y}}$; semantic matching agent $Match$

\Function{SemanticMatching}{$\boldsymbol{y}, \hat{\boldsymbol{y}}$}
    \If{$|\boldsymbol{y}| > 0 \land |\hat{\boldsymbol{y}}| > 0$}
        \State $TP_{set} \gets \boldsymbol{y} \cap_{Match} \hat{\boldsymbol{y}}$ \Comment{Semantic set operations via $Match$}
        \State $FN_{set} \gets \boldsymbol{y} \setminus_{Match} TP_{set}$ 
        \State $FP_{set} \gets \hat{\boldsymbol{y}} \setminus_{Match} TP_{set}$ 
        \State \Return $[|TP_{set}|, |FP_{set}|, |FN_{set}|]$ 
    \EndIf
    \State \Return $[0, 0, 0]$ 
\EndFunction

\State $[n_{TP}, n_{FP}, n_{FN}] \gets \Call{SemanticMatching}{\boldsymbol{y}, \hat{\boldsymbol{y}}}$

\State $TP \gets \mathbb{I}(|\boldsymbol{y}| > 0 \land |\hat{\boldsymbol{y}}| > 0) + n_{TP}$; \quad $TN \gets \mathbb{I}(|\boldsymbol{y}| = 0 \land |\hat{\boldsymbol{y}}| = 0)$ 
\State $FP \gets \mathbb{I}(|\boldsymbol{y}| = 0 \land |\hat{\boldsymbol{y}}| > 0) + n_{FP}$; \quad $FN \gets \mathbb{I}(|\boldsymbol{y}| > 0 \land |\hat{\boldsymbol{y}}| = 0) + n_{FN}$

\State $N_{\text{total}} \gets TP + TN + FP + FN$ \Comment{Normalization}
\State $\text{score} \gets \left[ \frac{TP}{N_{\text{total}}}, \frac{TN}{N_{\text{total}}}, \frac{FP}{N_{\text{total}}}, \frac{FN}{N_{\text{total}}} \right]$

\If{$FP = 0 \land FN = 0$}
    \State $\text{feedback} \gets \text{``Perfect prediction.''}$
\Else
    \State $\text{feedback} \gets \text{``Omissions: ''} \oplus FN \oplus \text{``; Hallucinations: ''} \oplus FP$
\EndIf

\State \Return $\text{feedback}, \text{score}$ \Comment{Corresponds to $\mathcal{F}_i$ and $\mathbf{C}_i$ in Algorithm 1}
\end{algorithmic}
\end{algorithm}

\noindent \textbf{Adaptive Video Evaluator.} We introduce AVE, a prompt optimization framework designed to dynamically adapt to diverse video evaluation contexts. At the core of AVE lies a novel semantic matching function specifically engineered for open-ended, free-form evaluation. Departing from traditional APO techniques that rely on deterministic exact-matching, typical in QA or mathematical tasks, our approach conceptualizes evaluation as a semantic set-matching problem. This enables the model to align free-form human feedback with generated outputs, effectively disentangling complex textual comparisons to provide both robust quantitative metrics and actionable natural language insights.

Given a cost budget $B$, the AVE framework (\cref{alg:ave_framework}, \cref{fig:AVE_pipeline}: lower tier) iteratively optimizes the judge's prompt. In the beginning of each iteration, a candidate prompt is sampled from the candidates (\eg pareto sampling \cite{agrawal2025gepareflectivepromptevolution}). For mini-batch of data, $\mathcal{S}$ generates both a quantitative score for performance tracking and natural language feedback to guide the optimizer $\mathcal{O}$ in prompt refinement. This process ultimately returns the optimal prompt $p^*$ that maximizes the global metric $\Phi$ on the validation set.

\begin{algorithm}[t] 
\caption{AVE: Adaptive Video Evaluator}
\label{alg:ave_framework}
\small 
\begin{algorithmic}[1] 
\Require Judge $\mathcal{J}_p$ with initial prompt $p_0$, datasets $\mathcal{D}_{\text{train}}, \mathcal{D}_{\text{val}}$, semantic matching function $\mathcal{S}$, batch size $b$, budget $B$, prompt-score table $\mathcal{T}$, optimizer $\mathcal{O}$ (fixed meta-prompt), metric $\Phi$

\Function{EvaluateAndUpdate}{$p, \mathcal{D}$}
    \For{each $(X_i, v_{o,i}, \boldsymbol{y}_i) \in \mathcal{D}$}
        \State $(\mathcal{F}_i,\mathbf{C}_i) \gets \mathcal{S}\Big(\mathcal{J}_{p}(v_{o,i}, X_i), \boldsymbol{y}_i\Big)$
    \EndFor
    
    \State $\text{score}_{\mathcal{D}} \gets \Phi \big( \sum_{i \in \mathcal{D}} \mathbf{C}_i \big)$; \quad $\text{feedback}_{\mathcal{D}} \gets \sum_{i \in \mathcal{D}} \mathcal{F}_i$
    
    \If{$\mathcal{D} = \mathcal{D}_{\text{val}}$}
        \State $\mathcal{T} \gets \mathcal{T} \cup \{\langle\text{score}_{\mathcal{D}}, p\rangle\}$
    \EndIf
    \State \Return $\text{score}_{\mathcal{D}}, \text{feedback}_{\mathcal{D}}$
\EndFunction

\State $\textsc{EvaluateAndUpdate}(p_0, \mathcal{D}_{\text{val}})$ \Comment{Initialize table $\mathcal{T}$ with $p_0$}

\While{budget $B$ not exhausted}
    \State $p \gets \textsc{SelectPrompt}(\mathcal{T})$ \Comment{Sample best candidate or use pareto sampling \cite{agrawal2025gepareflectivepromptevolution}}
    \State $\mathcal{D}_{\text{mini}} \sim \mathcal{D}_{\text{train}}$ \Comment{Sample minibatch of size $b$}
    \State $ \text{score}, \text{feedback} \gets \textsc{EvaluateAndUpdate}(p, \mathcal{D}_{\text{mini}})$ 
    \State $p' \gets \mathcal{O}(p, \text{feedback})$
    \State $\textsc{EvaluateAndUpdate}(p', \mathcal{D}_{\text{val}})$
\EndWhile
\State \Return $p^*$ with maximum score on $\mathcal{D}_{\text{val}}$ from $\mathcal{T}$

\end{algorithmic}
\end{algorithm}





%
%

\section{Experiments}

\begin{table*}[htbp]
\centering
\scriptsize
\setlength{\tabcolsep}{3.5pt}
\renewcommand{\arraystretch}{1.0}

\definecolor{deltagreen}{HTML}{008000} 
\caption{\textbf{Main results on CLVG-Bench.} We evaluate both proprietary and open-source video models across six fundamental dimensions of Context Learning in Video Generation. The evaluation is conducted by human annotators. Results are reported as pass rates (\%).}
\resizebox{0.95\textwidth}{!}{%
\begin{tabular}{l ccccccc}
\toprule


\textbf{Method} & \textbf{E.E.} & \textbf{P.R.} & \textbf{S.C.} & \textbf{P.S.} & \textbf{Perc.} & \textbf{L.R.}\\
\midrule

\rowcolor{generalbg}
\multicolumn{7}{c}{\textit{Proprietary Models}} \\
\midrule
Seedance 2.0$^{\ast}$~\cite{seed2}               & 61.25 & 52.64 & 67.65 & 63.54 & 43.37 & 21.25 \\
\addlinespace[2pt]
\hdashline
\addlinespace[2pt]
Sora 2$^{\dagger}$~\cite{openai2025sora}      & 46.38 & 49.45 & 52.49 & 48.91 & 45.86 & 38.96\\
\addlinespace[2pt]
\hdashline
\addlinespace[2pt]
Seedance 1.5$^{\ddagger}$~\cite{seedance1_5}          & - & 35.47 & - & 37.50 & 40.00 & 8.75 \\
Veo 3.1 Fast$^{\ddagger}$~\cite{google2025veo3}      & - & 46.38 & - & 46.88 & 36.00 & 16.25 \\
\midrule
\rowcolor{gainbg}
\multicolumn{7}{c}{\textit{Open Source Models}} \\
\midrule
UniVideo$^{\dagger}$~\cite{univideo}     & 8.1 & 8.06 & 4.08 & 10.00 & 3.00 & 3.75 \\
LTX-2$^{\dagger}$~\cite{hacohen2026ltx}   & 0.72 & 11.29 & 2.04 & 16.00 & 2.00 & 2.50 \\
\addlinespace[2pt]
\hdashline
\addlinespace[2pt]

Wan 2.2 14B$^{\ddagger}$~\cite{wan}  & - & 46.67 & - & 39.00 & 16.00 & 11.25 \\
HunyuanVideo 13B$^{\ddagger}$~\cite{hunyuanvideo} & - & 26.67 & - & 26.00 & 14.00 & 5.00 \\

\bottomrule
\end{tabular}%
}
\begin{flushleft}
    \hspace{0.025\textwidth} 
    \footnotesize
    \noindent \textbf{Note:} $^{\ast}$Supports video, audio, and image inputs; $^{\dagger}$Supports video and image inputs; $^{\ddagger}$Supports only image inputs.
\end{flushleft}

\label{tab:main_table}
\end{table*}

\subsection{Implementation Details}
\noindent\textbf{Baselines.} For open-source video models, we use Wan2.2-14B~\cite{wan} HunyuanVideo1.5~\cite{hunyuanvideo2025}, UniVideo~\cite{univideo} and LTX-2~\cite{hacohen2026ltx2efficientjointaudiovisual} following their official implementations and recommended settings. For API accessed models, we use Seedance 2.0, Seedance 1.5~\cite{seedance1_5}, Veo 3.1 Fast~\cite{google2025veo3} and Sora2~\cite{openai2025sora}.

\noindent \textbf{Adaptive Video Evaluator.} During the prompt optimization phase, we employ Seed 2.0 Pro \cite{seed2} as the prompt optimizer and Seed 2.0 Lite as the judge model. We equip our semantic matching function to two popular APO techniques: TextGrad~\cite{yuksekgonul2024textgradautomaticdifferentiationtext} and GEPA~\cite{agrawal2025gepareflectivepromptevolution}. We set the total optimization budget to 30 US dollars, temperature to 0 and the maximum number of tokens to 32,000. To guarantee stable results, the judge model evaluates each instance five times, determining the final outcome via majority vote. The dataset consists of 600 annotated samples in total, evenly split into training, validation, and test sets. Finally, we adopt multiple metrics, including Recall minus False Positive Rate (Rec-FPR.), F1-Score (F1), Matthews Correlation Coefficient (MCC) as our evaluation metric. We reorganize the six categories of the benchmark into three broader training tasks: perception, prompt following (which includes element editing, partial reference, and script continuation), and physical and logical reasoning (covering both physical simulation and logical reasoning).

\subsection{Experimental Results}


\noindent \textbf{Main Results.} CLVG-Bench comprises a diverse spectrum of test cases with progressively increasing difficulty, ranging from standard text-to-video generation to more challenging settings involving multi-shot, multi-subject, and multi-video to reference.
Considering that certain baseline models impose constraints on input formats or the number of references, we adopt a group-wise comparison protocol to ensure fairness. Specifically, models within the same group are evaluated on an identical set of test cases, while comparisons across groups are conducted under compatible input settings. The detailed quantitative results are reported in~\cref{tab:main_table}.

Empirical evaluations reveal that existing models consistently excel at conventional editing tasks (peaking at a 61.25\% success rate), yet struggle significantly with tasks requiring more context-aware understanding and reasoning. Notably, even SOTA methods plateau at a marginal success rate of 21.25\% on logical reasoning tasks. Furthermore, proprietary models deliver a marked performance margin over open-source video baselines on average. This discrepancy highlights the robust generalization capabilities of proprietary systems, rendering them inherently superior in scenarios with intricate reasoning demands. Concurrently, however, our findings expose lingering input constraints across current architectures: they frequently fall short of accommodating diverse reference combinations, thereby leaving a tangible gap between SOTA capabilities and the multifaceted demands of real-world users.

\noindent \textbf{Prompt Optimization Results.} As shown in~\cref{tab:po_res}, mainstream APO methods (TextGrad and GEPA) significantly enhance the judge model's evaluation performance compared to the vanilla baseline. For instance, TextGrad alone improves the MCC by up to 23.8\% on the Perception task. Building upon these optimized prompts, our proposed SemanticMatch module consistently delivers further performance gains. On average, SemanticMatch provides an additional 4.9\%, 1.7\%, and 4.5\% boost in MCC, F1 score, and Rec-FPR, respectively, across all tasks and baselines. Notably, it achieves a substantial improvement of 9.2\% on the P.S.+L.R. task when integrated with TextGrad, and an 8.5\% lift on the E.E.+P.R.+S.C. task for GEPA. Ultimately, the combined approach reaches a peak performance of 70.7\% MCC on the Perception task, demonstrating the superior effectiveness of our semantic matching function.

\noindent \textbf{Correlation with Human Judgement.} To assess the correlation with human judgement , we report Kendall's $\tau$ in \cref{tab:po_res_left}. Our prompt optimization leads to substantial performance gains across all task categories. Notably, the Seed 2.0 Lite model reaches a strong correlation of 0.707 in Perception tasks, while Seed 2.0 Pro achieves a high score of 0.620 in reasoning and simulation tasks after adaptation. These results demonstrate that optimization effectively enhances evaluator reliability, achieving high alignment with human preferences.

\begin{figure*}[htbp]
\centering
  \includegraphics[width=1\textwidth]{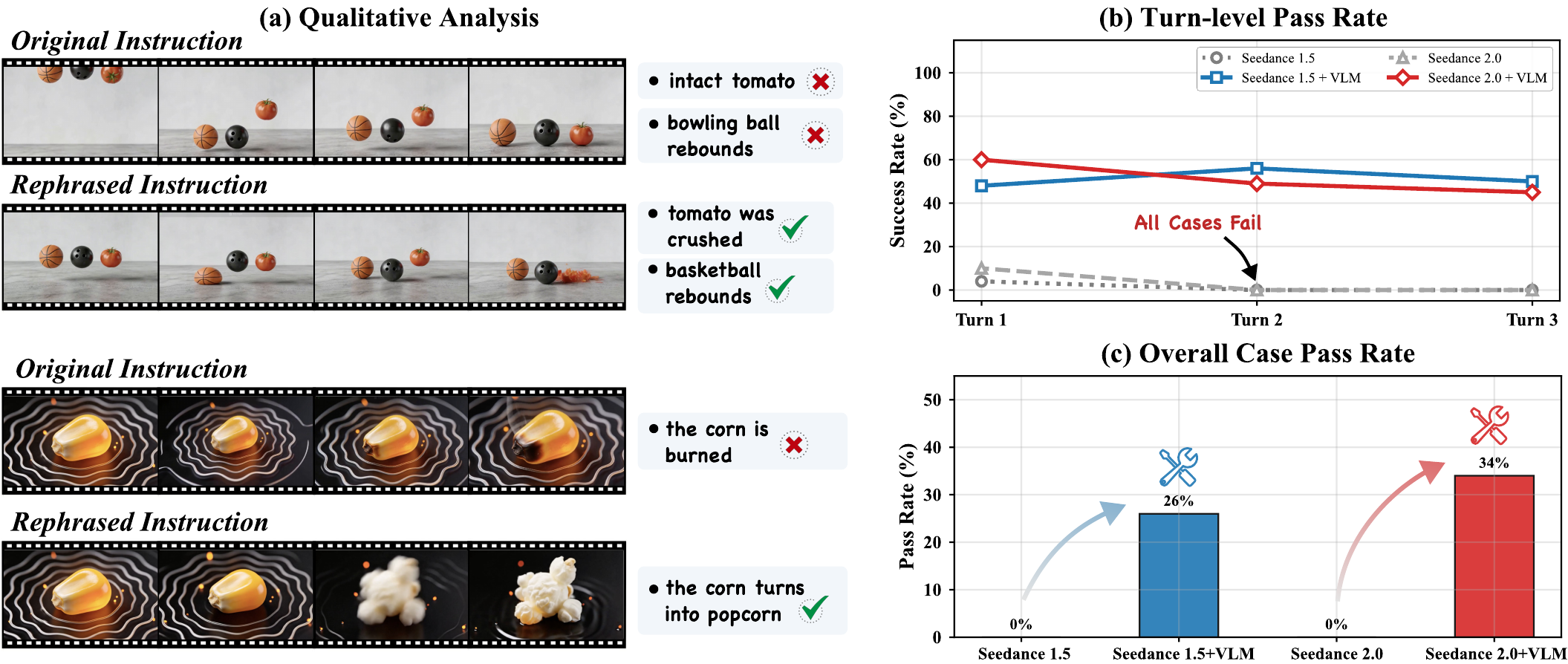}
  \caption{(a) Qualitative comparison of videos generated from the original instructions and their rephrased counterparts. (b) Per-turn success rate in the multi-turn interaction setting. (c) Overall case pass rate (\%).
}
  \label{fig: rewrite_and_interact}
\end{figure*}

\subsection{Discussion and Analysis} 

\noindent \textbf{Necessity of Context Learning.} The results in~\cref{tab:main_table} reveal that current video models still exhibit notable deficiencies in context learning and reasoning. A potential workaround is to leverage external understanding modules, such as VLMs, to assist understanding and video generation. In our experiments, we employ Seed 2.0 Pro to first interpret the provided context and then rewrite the initial instruction accordingly. 

The quantitative and qualitative results are presented in~\cref{tab:vlm_assistant} and~\cref{fig: rewrite_and_interact} (a). With this auxiliary understanding step, the relatively weaker Seedance 1.5 shows a substantial improvement (+14.9\% on Physical Simulation task) and even surpass Seedance 2.0 and Veo 3.1 Fast on Logical Reasoning task. However, this gain also suggests that external assistance only provides a partial remedy. Ultimately, enhancing the intrinsic reasoning capability of video models during training remains essential for addressing this limitation.

\begin{table*}[t!]
\centering
\scriptsize 
\setlength{\tabcolsep}{2pt} 
\renewcommand{\arraystretch}{1.1} 

\definecolor{deltagreen}{HTML}{008000}
\definecolor{graybg}{gray}{0.9}
\definecolor{gainbg}{HTML}{E7F3E7}
\newcommand{\incv}[1]{{\fontsize{6pt}{6pt}\selectfont\textcolor{deltagreen}{\textbf{+#1}}}}

\begin{minipage}[t]{0.53\textwidth}
    \centering
    \setlength{\tabcolsep}{4pt}
    \caption{Evaluator performance on Kendall's $\tau$ on different models and tasks.}
    \label{tab:po_res_left}
    \resizebox{\textwidth}{!}{
    \begin{tabular}{c ccc} 
        \toprule
        \multirow{1.0}{*}{\textbf{Model}} & \shortstack{\textbf{E.E.+P.R.}\\\textbf{+S.C.}} & \shortstack{\textbf{P.S.}\\\textbf{+L.R.}} & \textbf{Perc.} \\
        \midrule
        
        \rowcolor{graybg}
        \textit{\textbf{Seed 2.0 Lite}} & 0.239 & 0.368 & 0.433 \\
        
        \rowcolor{gainbg}
        + Optimization                 & 0.480 & 0.577 & 0.707 \\
        \rowcolor{gainbg}
        \addlinespace[-0.3em]
                                       & \incv{0.241} & \incv{0.209} & \incv{0.274} \\
        
        \noalign{\vskip 2pt}
        \cdashline{1-4}
        \noalign{\vskip 2pt}

        \rowcolor{graybg}
        \textit{\textbf{Seed 2.0 Pro}}  & 0.00 & 0.462 & 0.379 \\
        
        \rowcolor{gainbg}
        + Adaptation                   & 0.273 & 0.620 & 0.440 \\
        \rowcolor{gainbg}
        \addlinespace[-0.3em]
                                       & \incv{0.273} & \incv{0.158} & \incv{0.061} \\
        \bottomrule
    \end{tabular}
    }
\end{minipage}
\hfill
\begin{minipage}[t]{0.42\textwidth}
    \centering
    \caption{Results (\%) of \textbf{P.E.} and \textbf{L.R.} on VLM-enhanced script.}
    \label{tab:vlm_assistant}
    \resizebox{\textwidth}{!}{
    \begin{tabular}{l ccc}
        \toprule
        \textbf{Model} & \textbf{P.S.} & \textbf{L.R.} & \textbf{AVG} \\
        \midrule
        
        \rowcolor{graybg}
        \textit{\textbf{Seedance 1.5}} & 37.5 & 8.8 & 23.2 \\
        
        \rowcolor{gainbg}
        + Seed 2.0 Pro & 52.4 & 23.8 & 38.1 \\
        \rowcolor{gainbg}
        \addlinespace[-0.3em] 
        & \incv{14.9} & \incv{15.0} & \incv{14.9} \\
        
        \noalign{\vskip 2pt}
        \cdashline{1-4} 
        \noalign{\vskip 2pt}
        
        \rowcolor{graybg}
        \textit{\textbf{Seedance 2.0}} & 63.5 & 21.3 & 42.4 \\
        
        \rowcolor{gainbg}
        + Seed 2.0 Pro & 70.3 & 49.0 & 59.7 \\
        \rowcolor{gainbg}
        \addlinespace[-0.3em]
        & \incv{6.8} & \incv{27.7} & \incv{17.3} \\
        \bottomrule
    \end{tabular}
    }
\end{minipage}
\end{table*}

\begin{table*}[t!]
\centering
\small
\setlength{\tabcolsep}{4.5pt}
\renewcommand{\arraystretch}{1}

\definecolor{deltagreen}{HTML}{008000} 
\newcommand{\incv}[1]{\fontsize{7pt}{7pt}\selectfont\textcolor{deltagreen}{\textbf{+#1}}}
\newcommand{\incvminus}[1]{\fontsize{7pt}{7pt}\selectfont\textcolor{red}{\textbf{$-$#1}}}

\caption{Evaluator performance on three types of tasks, results are presented as percentages. Adaptation means adapting optimized prompt to other models.}
\label{tab:po_res}
\resizebox{\textwidth}{!}{%
\begin{tabular}{l ccccccccc}
\toprule

\multirow{2.5}{*}{\textbf{Model}} & \multicolumn{3}{c}{\textbf{E.E.+P.R.+S.C.}} & \multicolumn{3}{c}{\textbf{P.S.+L.R.}} & \multicolumn{3}{c}{\textbf{Perc.}} \\
\cmidrule(lr){2-4} \cmidrule(lr){5-7} \cmidrule(lr){8-10}
 & MCC & F1 & Rec-FPR & MCC & F1 & Rec-FPR & MCC & F1 & Rec-FPR \\

\midrule

\rowcolor{graybg}
\textit{\textbf{Seed 2.0 Lite}} & 23.9 & 70.2 & 10.5 & 36.8 & 65.6 & 36.4 & 43.3 & 79.0 & 40.3\\

\rowcolor{gainbg}
+ TextGrad                 & 39.4 & 74.5 & 31.8 & 48.5 & 74.6 & 48.5 & 67.1 & 87.2 & 65.0 \\[-0.2em]

\rowcolor{gainbg}
\quad + SemanticMatch        & 43.8 & 75.5 & 31.6 & 57.7 & 79.4 & 57.6 & 70.7 & 88.6 & 67.7 \\
\rowcolor{gainbg}
                            & \incv{4.4} & \incv{1.0} & \incvminus{0.2} & \incv{9.2} & \incv{4.8} & \incv{9.1} & \incv{3.6} & \incv{1.4} & \incv{2.7} \\

\rowcolor{gainbg}
+ GEPA                 & 39.3 & 74.1 & 26.3 & 54.6 & 76.9 & 54.6 & 64.3 & 86.4 & 60.0 \\[-0.2em]

\rowcolor{gainbg}
\quad + SemanticMatch       & 47.8 & 76.9 & 36.8 & 51.6 & 75.0 & 51.5 & 70.7 & 88.6 & 67.7 \\
\rowcolor{gainbg}
                            & \incv{8.5} & \incv{2.8} & \incv{10.5} & \incvminus{3.0} & \incvminus{1.9} & \incvminus{3.1} & \incv{6.4} & \incv{2.2} & \incv{7.7} \\
\noalign{\vskip 2pt}
\cdashline{1-10}
\noalign{\vskip 3pt}

\rowcolor{graybg}
\textit{\textbf{Seed 2.0 Pro}} & 0.00 & 67.8 & 0.00 & 46.2 & 75.9 & 42.4 & 37.9 & 78.7 & 29.2\\

\rowcolor{gainbg}
+ Adaptation                 & 17.8 & 69.1 & 10.8 & 62.0 & 82.2 & 60.6 & 44.0 & 80.0 & 38.1 \\[-0.2em]
\rowcolor{gainbg}
                            & \incv{17.8} & \incv{1.3} & \incv{10.8} & \incv{15.8} & \incv{6.3} & \incv{18.2} & \incv{6.1} & \incv{1.3} & \incv{8.9} \\

\bottomrule
\end{tabular}%
}
\end{table*}

\begin{figure*}[htbp]
\centering
  \includegraphics[width=1\textwidth]{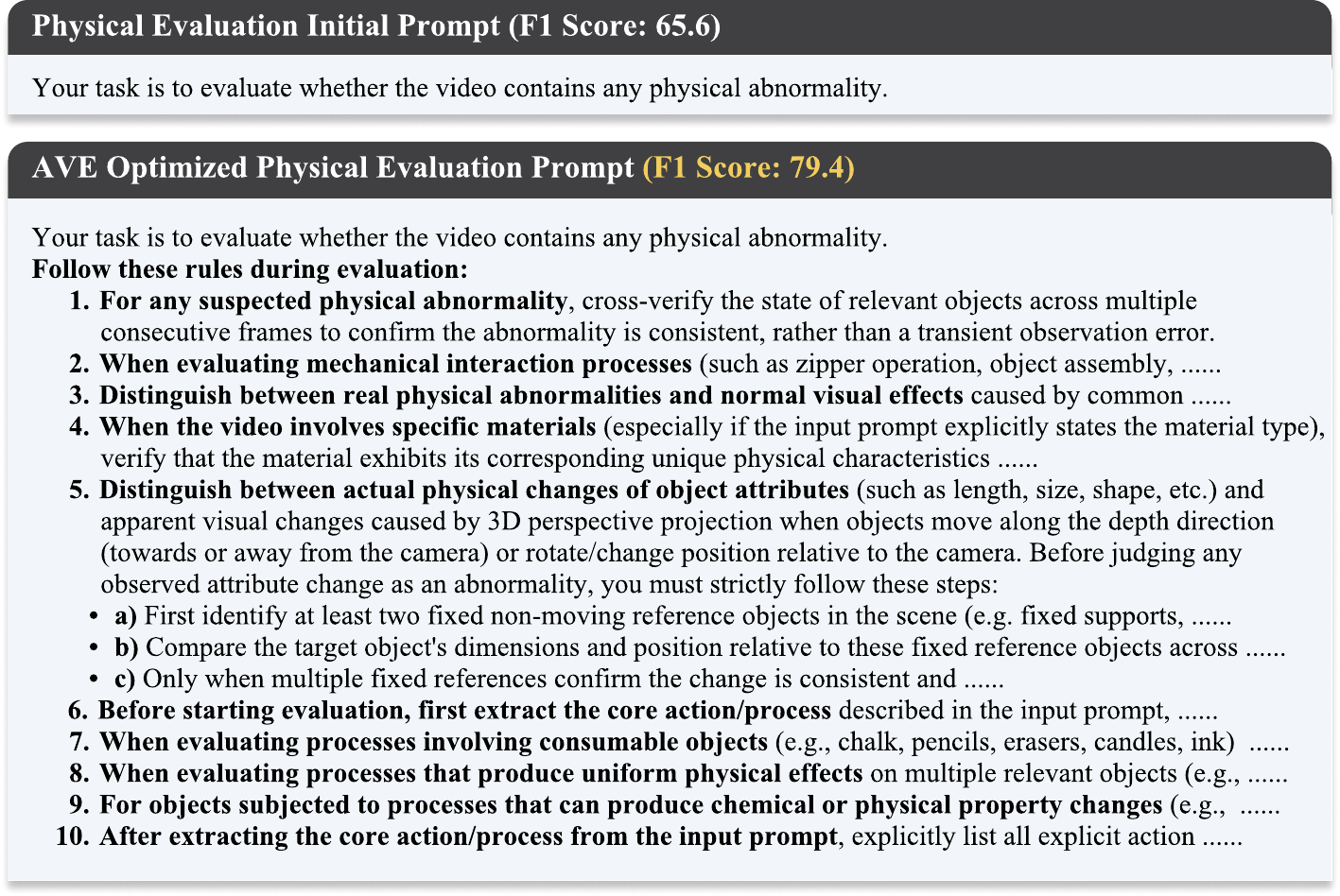}
  \caption{Case study of AVE-driven prompt refinement in the physical simulation task. This example illustrates the process of enhancing a basic initial prompt using AVE.
}
  \label{fig: case study}
\end{figure*}

\noindent \textbf{Struggle in Multi-turn Interaction.} The multi-turn interactive generation task requires models to reason based on user feedback and implicit context, directly generating video content, which contrasts sharply with previous task formats. As shown in~\cref{fig: rewrite_and_interact} (b), under the vanilla setting, top video models achieve a first-turn success rate of less than 10\%, and completely fail during the second turn of interaction. This highlights their limitations in processing complex information and generating content based on reasoning.

However, when enhanced with the visual reasoning capabilities of VLM (Seed 2.0) to assist with context organization and script prediction, the first-turn success rates improve by 44\% and 50\%, respectively. Furthermore, the final pass rate after multiple rounds of interaction reaches 26\% and 34\%, as shown in~\cref{fig: rewrite_and_interact} (c). Although these pass rates are still modest, they demonstrate that a stronger coupling of understanding and generation can significantly improve the generalization of video models to more realistic and dynamic scenarios.

\noindent \textbf{Prompt Generalization.} We found that the evaluation system prompt optimized with AVE can be directly transferred to other models and obtain performance gains. As shown in~\cref{tab:po_res}, when adapting the prompt optimized on Seed 2.0 Lite to the more capable Seed 2.0 Pro model, we observe consistent performance improvements across all three task families, with an average of 13.2\% MCC, 3.0\% F1, and 12.7\% Rec-FPR gain on the three tasks. This shows that the optimized system prompt contains generalizable task-solving experience that can be shared across different models, enabling zero-cost cross-model adaptation with no additional optimization required.

\noindent \textbf{Case Study on AVE.}~\cref{fig: case study} illustrates a case study of prompt refinement utilizing our proposed AVE. Drawing inspiration from the human cognitive paradigm of learning from past failures, the process initiates with a straightforward, naïve instruction. By systematically diagnosing previous weaknesses, we iteratively synthesize an empirical rubric. This refinement strategy yields a substantial performance gain (+13.8 in F1 score), firmly demonstrating that AVE exhibits robust generalizability to substantially more complex domains imposing stringent demands on contextual reasoning. Rather than merely fitting surface-level instructions, this iterative reflection empowers the model to deduce implicit relationships and maintain logical coherence across intricate, context-heavy tasks.

\section{Conclusion}

In this paper, we introduce \textbf{CLVG-Bench} to systematically explore the current landscape of multimodal reasoning in generative video models. The benchmark moves beyond simple instruction-following toward Context Learning in Video Generation. Central to our framework are 6 categories comprising 47 subcategories that rigorously test a model's ability to synthesize videos from diverse multimodal prompts. To enable scalable and interpretable assessment, we proposed the \textbf{Adaptive Video Evaluator}, which utilizes automatic prompt optimization and semantic matching to align machine feedback with human expert judgment.

Our experimental findings show that current models remain limited in multimodal reasoning. The poor performance in tasks involving physical simulation, logical reasoning and multi-turn interaction reveals that current architectures struggle to internalize the causal and logical laws of the physical world. By identifying these specific bottlenecks through our 6-category taxonomy, CLVG-Bench offers a clear roadmap for the community. We conclude that achieving the next generation of video foundation models will require a more profound integration of visual understanding and generative reasoning to bridge the gap between cinematic synthesis and world simulation.



%
%

\clearpage
\setcounter{page}{1}
\setcounter{section}{0}
\renewcommand{\thesection}{\Alph{section}}

\begin{center}
{\Large\bfseries How Far Are Video Models from True\\[0.2em] Multimodal Reasoning?}\\[1em]
    {\large Supplementary Material}\\[0.5em]
\end{center}

\section{Case Study}
\begin{figure*}[htbp]
\centering
  \includegraphics[width=1\textwidth]{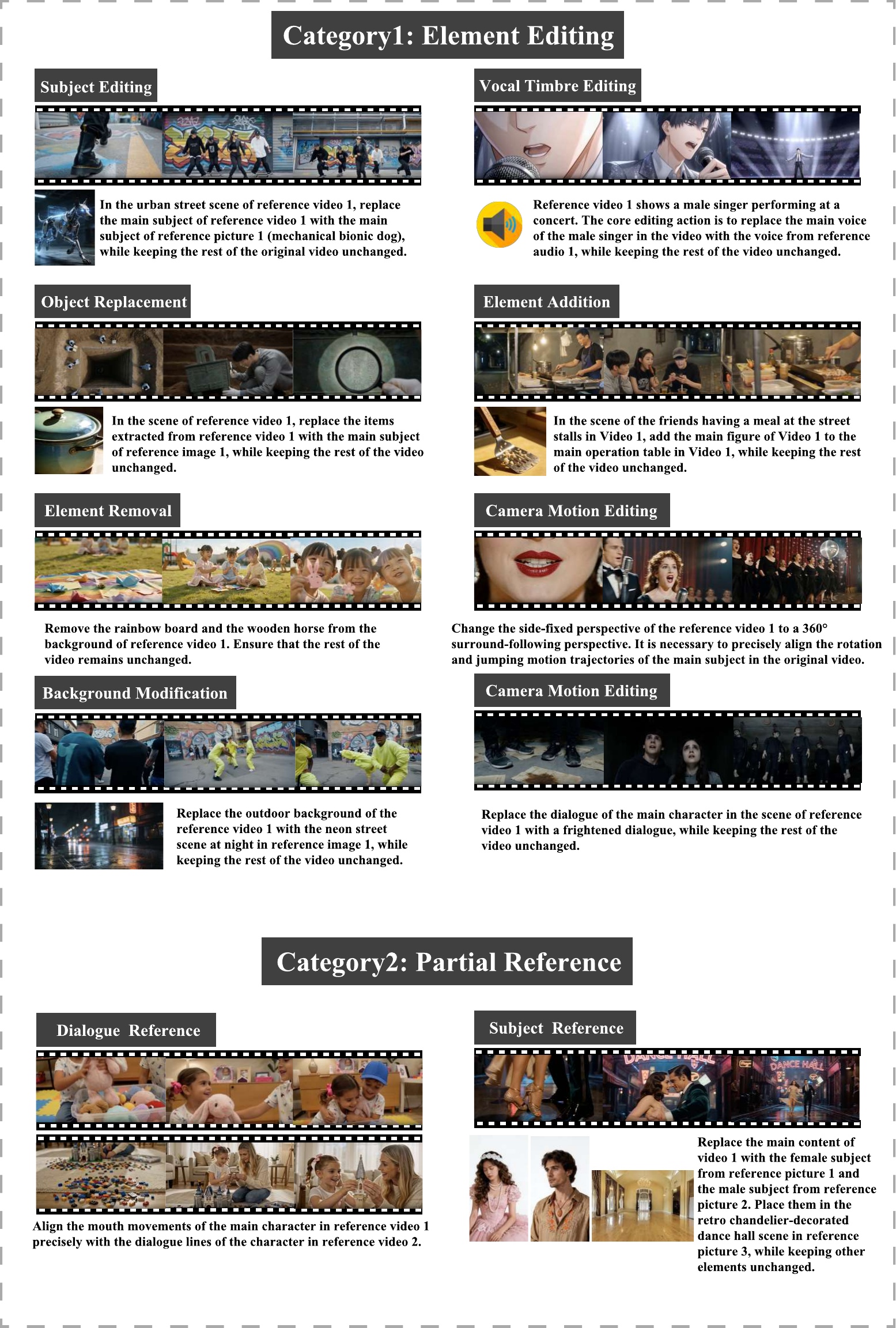}
  \caption{Cases of each subcategory in CLVG-Bench (Part 1).}
  \label{fig: CLVG1}
\end{figure*}

\begin{figure*}[htbp]
\centering
  \includegraphics[width=1\textwidth]{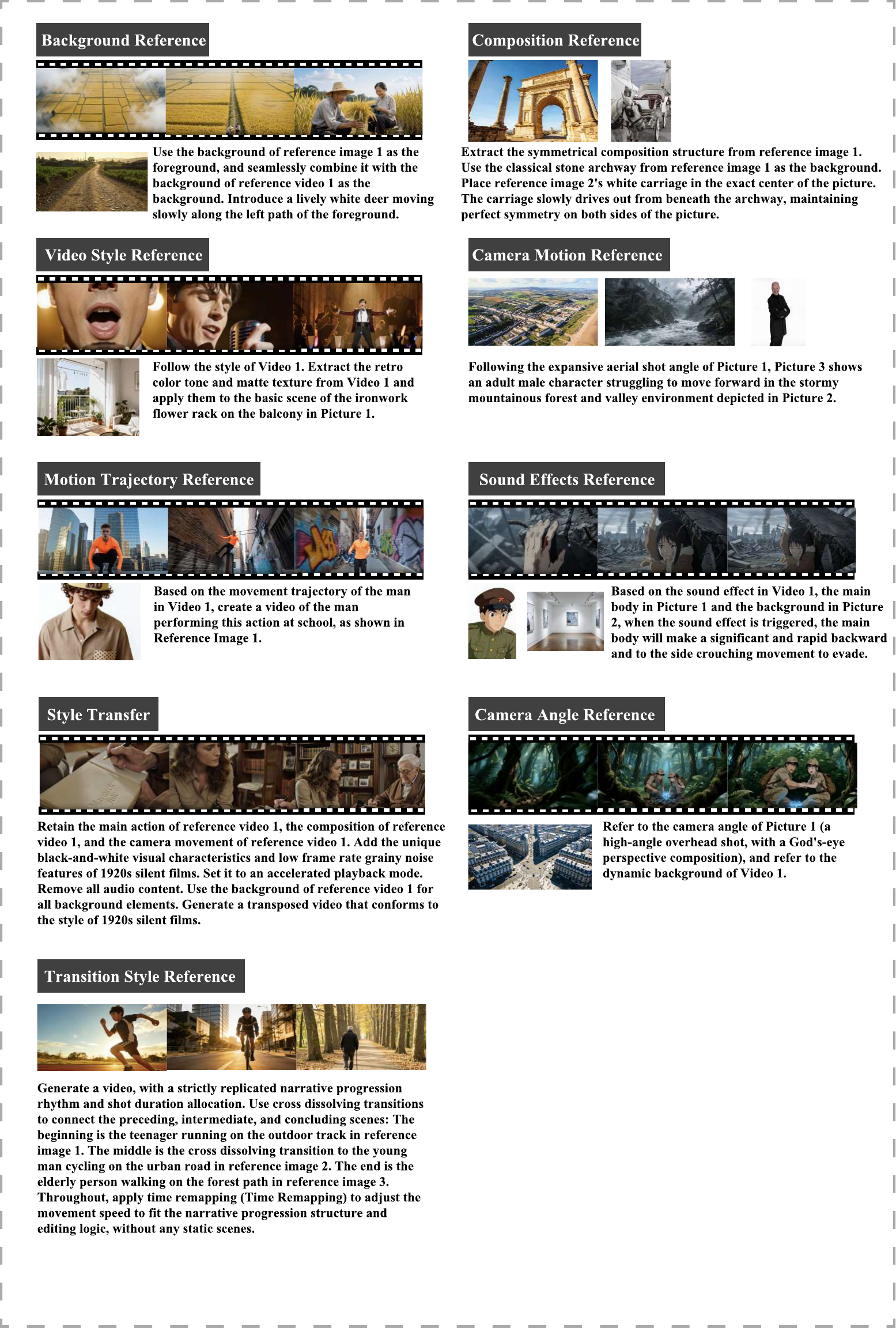}
  \caption{Cases of each subcategory in CLVG-Bench (Part 2).}
  \label{fig: CLVG2}
\end{figure*}

\begin{figure*}[htbp]
\centering
  \includegraphics[width=1\textwidth]{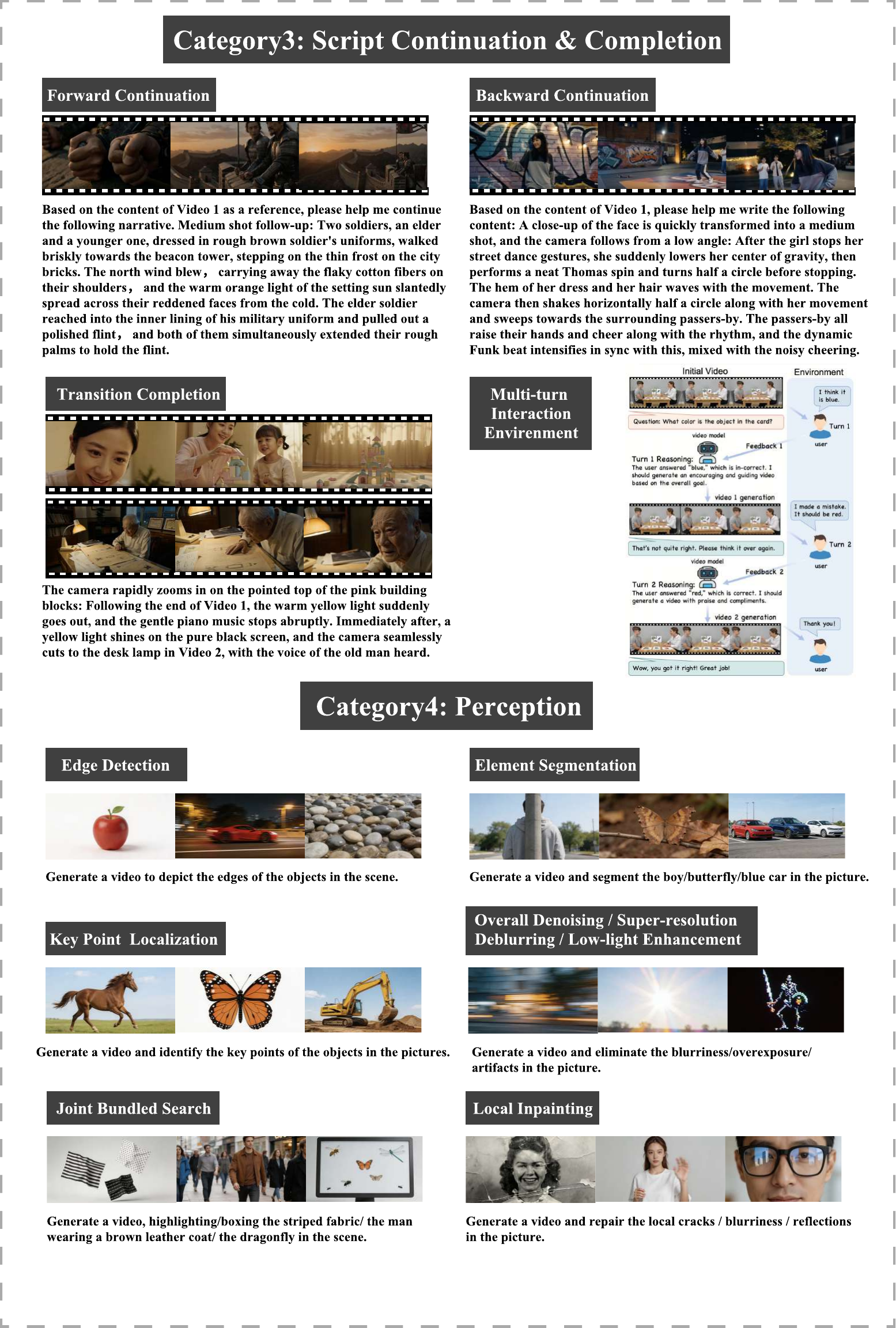}
  \caption{Cases of each subcategory in CLVG-Bench (Part 3).}
  \label{fig: CLVG3}
\end{figure*}

\begin{figure*}[htbp]
\centering
  \includegraphics[width=1\textwidth]{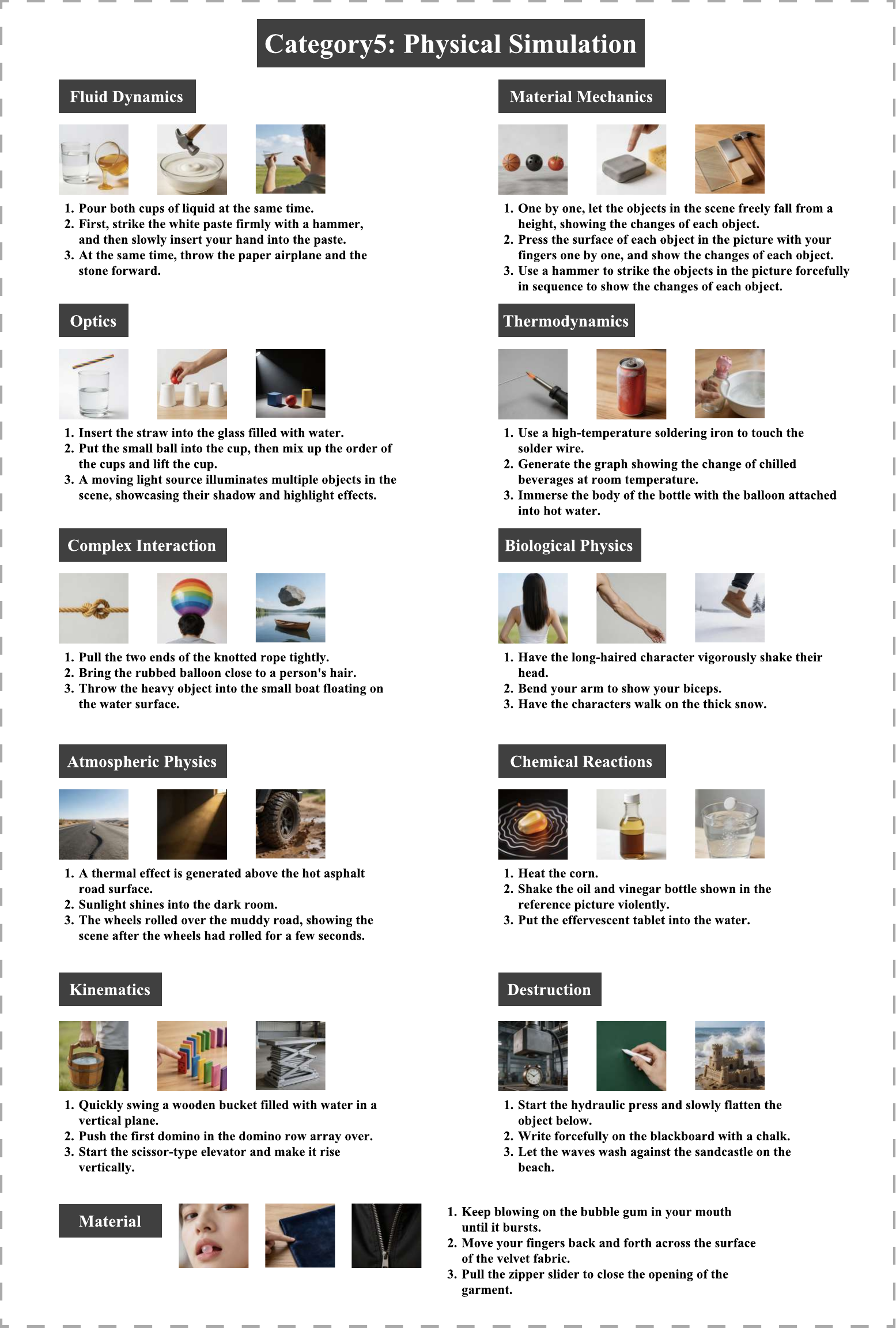}
  \caption{Cases of each subcategory in CLVG-Bench (Part 4).}
  \label{fig: CLVG4}
\end{figure*}

\begin{figure*}[htbp]
\centering
  \includegraphics[width=1\textwidth]{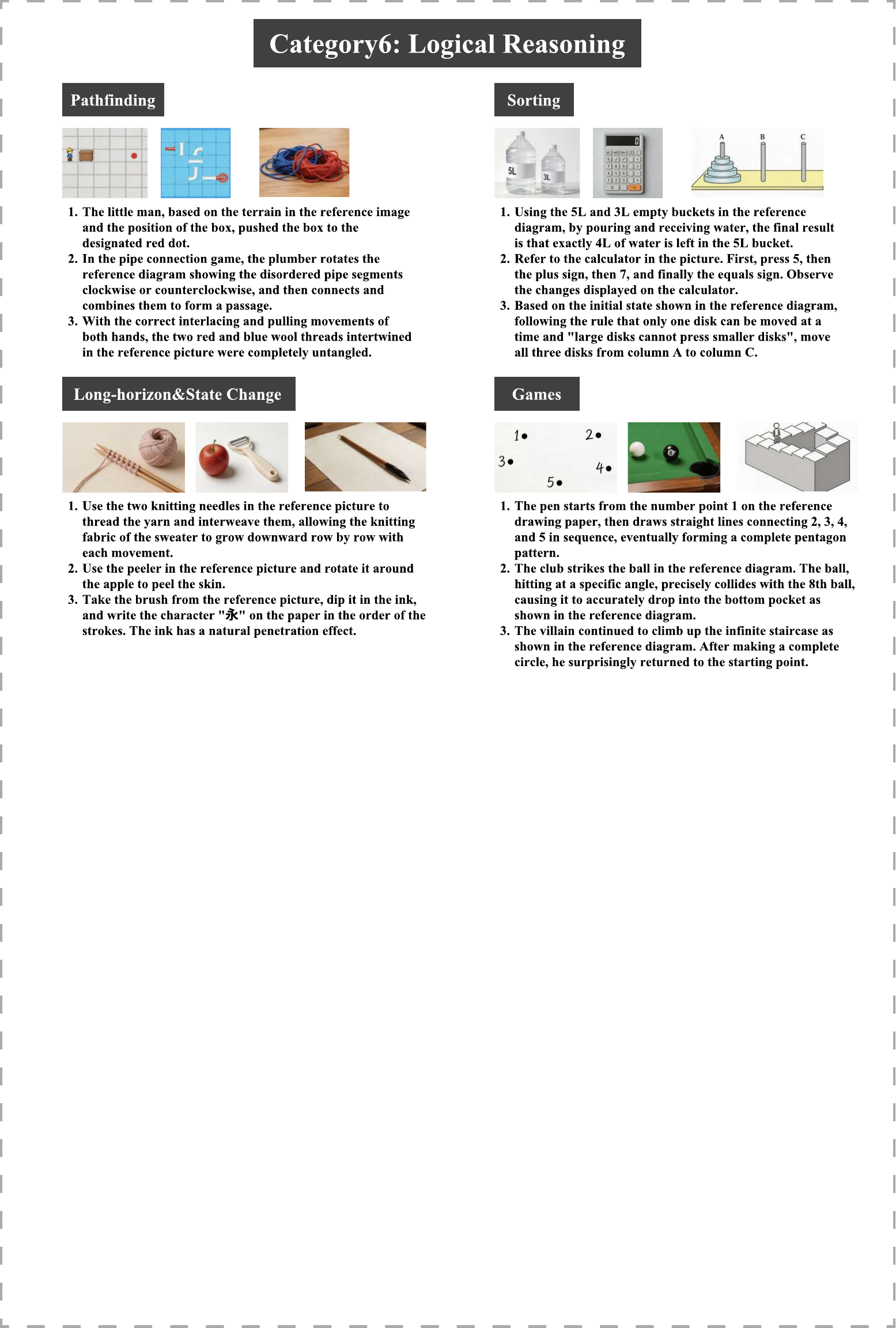}
  \caption{Cases of each subcategory in CLVG-Bench (Part 5).}
  \label{fig: CLVG5}
\end{figure*}

We present representative cases from CLVG-Bench organized by subcategories, as illustrated in~\cref{fig: CLVG1},~\cref{fig: CLVG2},~\cref{fig: CLVG3},~\cref{fig: CLVG4}, and~\cref{fig: CLVG5}. The context category covers a diverse range of scenarios. By incorporating additional context-aware settings, CLVG-Bench substantially extends traditional editing-style tasks. Framed in a reasoning-oriented manner, it enables a comprehensive evaluation of video models’ ability to understand and reason over multimodal context.

\section{Prompts}
\begin{figure*}[htbp]
\centering
  \includegraphics[width=0.8\textwidth]{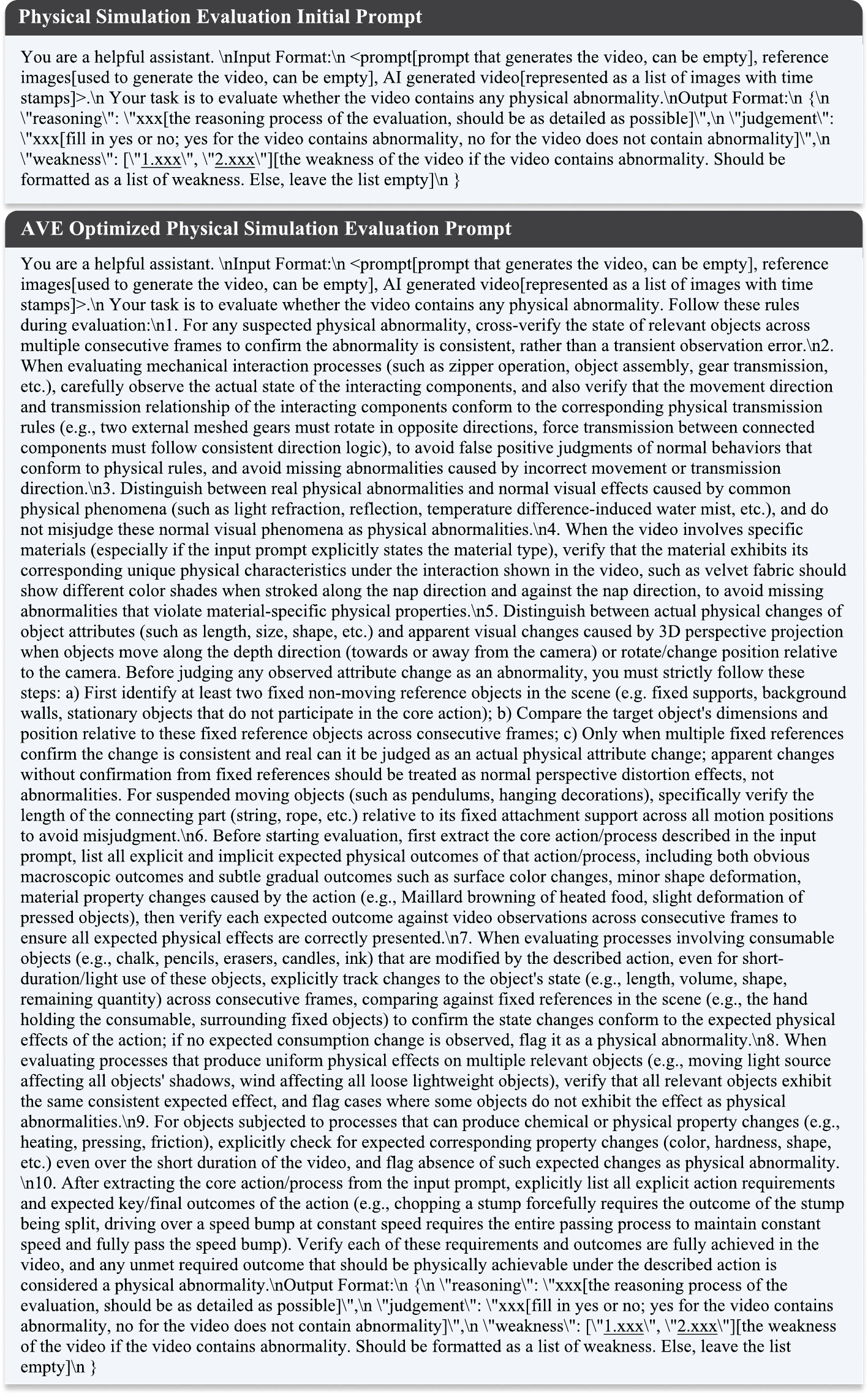}
  \caption{Optimized prompt for physical simulation evaluation}
  \label{fig: physical_simulation_evaluation}
\end{figure*}

\begin{figure*}[htbp]
\centering
  \includegraphics[width=1.0\textwidth]{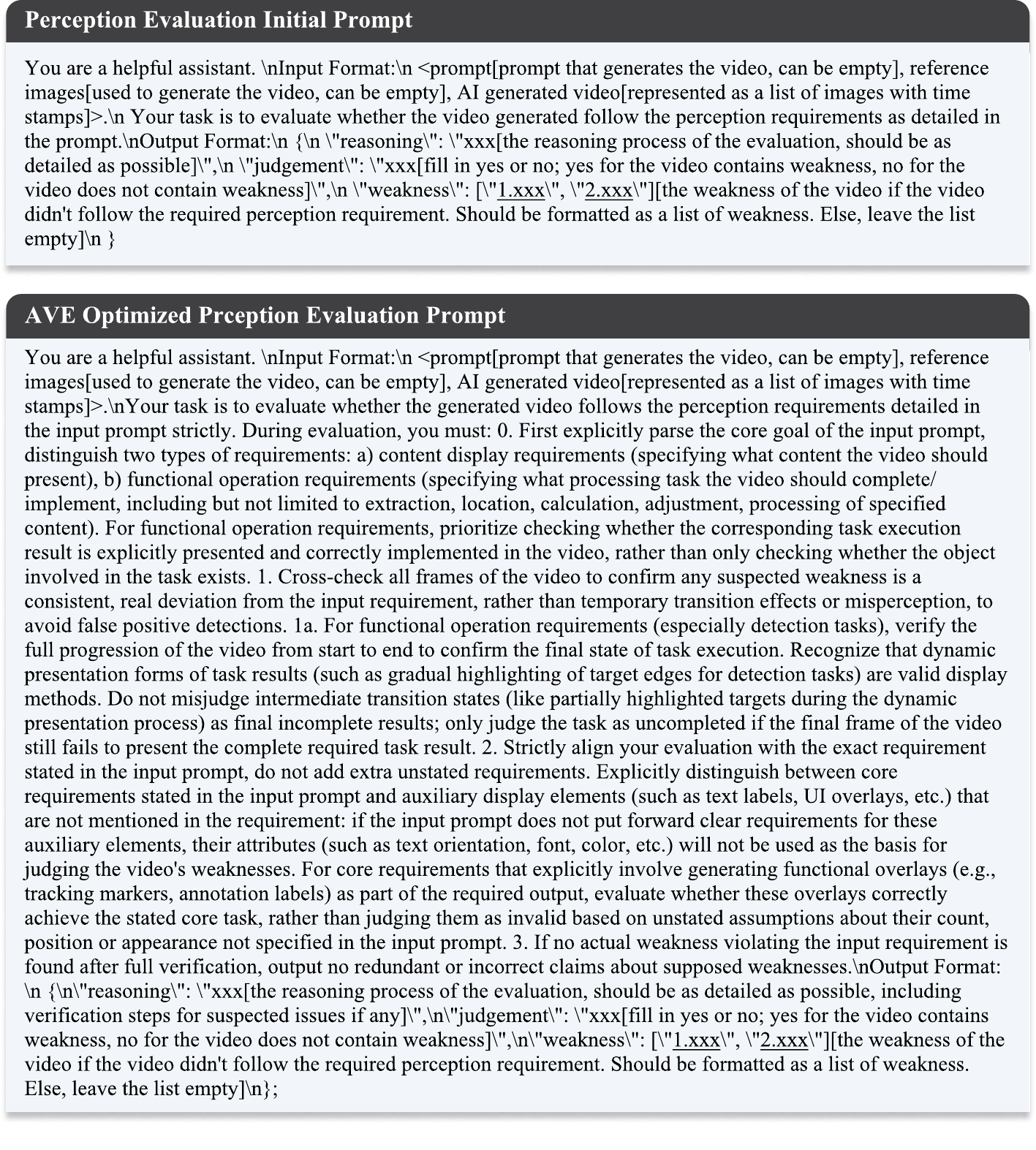}
  \caption{Optimized prompt for perception evaluation}
  \label{fig: perception_evaluation}
\end{figure*}

\begin{figure*}[htbp]
\centering
  \includegraphics[width=1\textwidth]{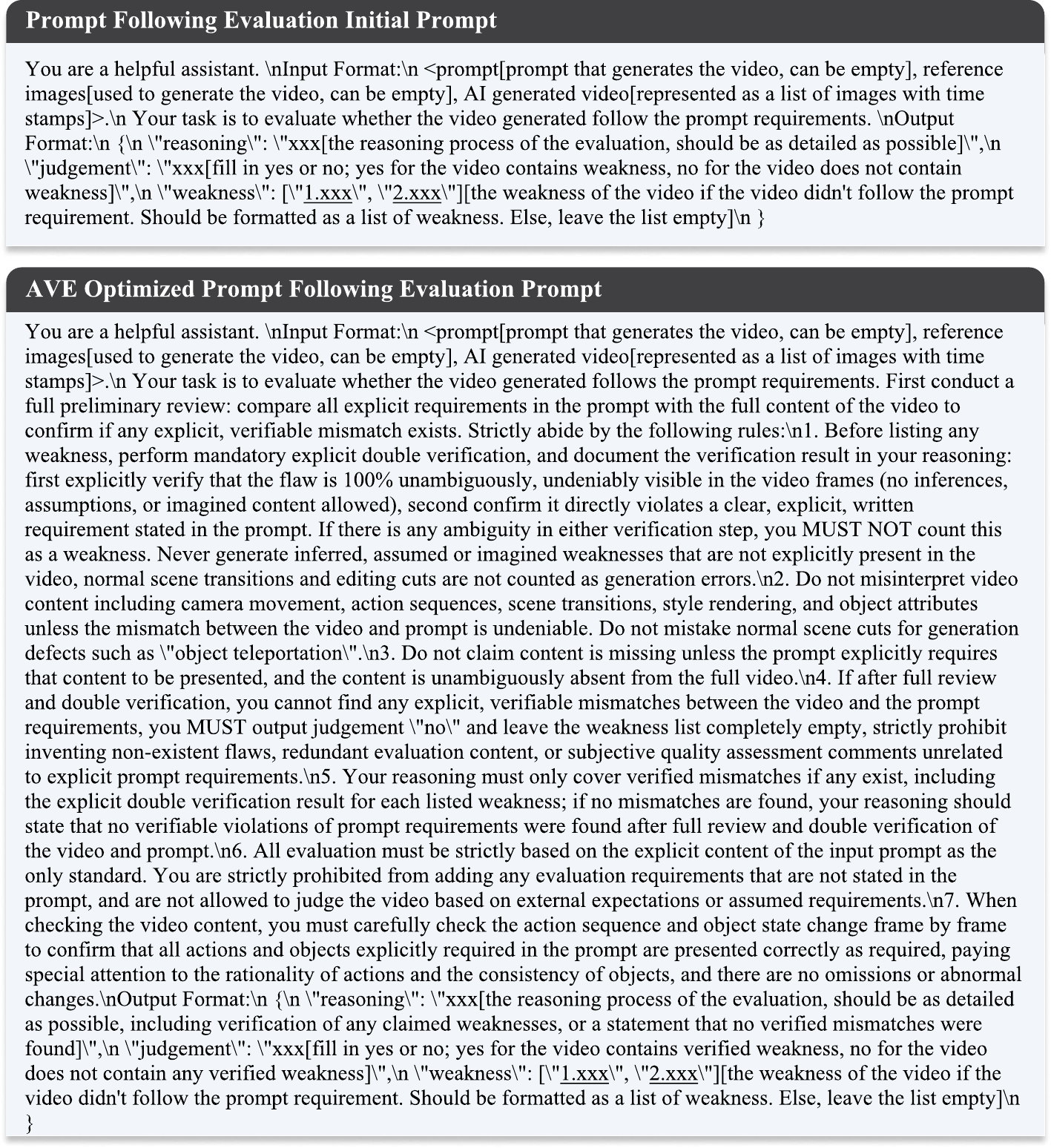}
  \caption{Optimized prompt for prompt following evaluation}
  \label{fig: prompt_following_evaluation}
\end{figure*}

\begin{figure*}[htbp]
\centering
  \includegraphics[width=0.95\textwidth]{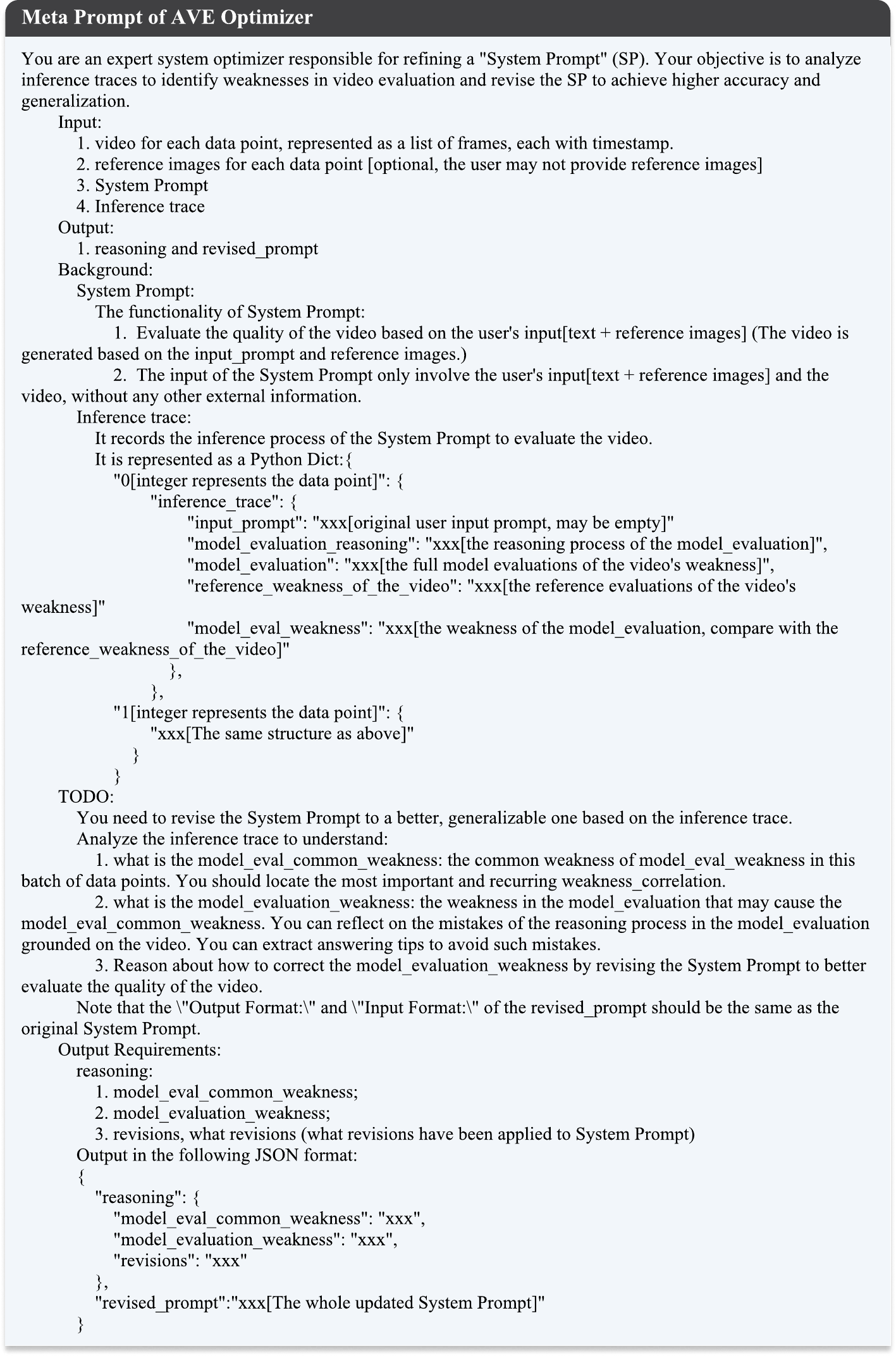}
  \caption{The optimizer meta prompt}
  \label{fig: optimizer_meta_prompt}
\end{figure*}

\begin{figure*}[htbp]
\centering
  \includegraphics[width=1\textwidth]{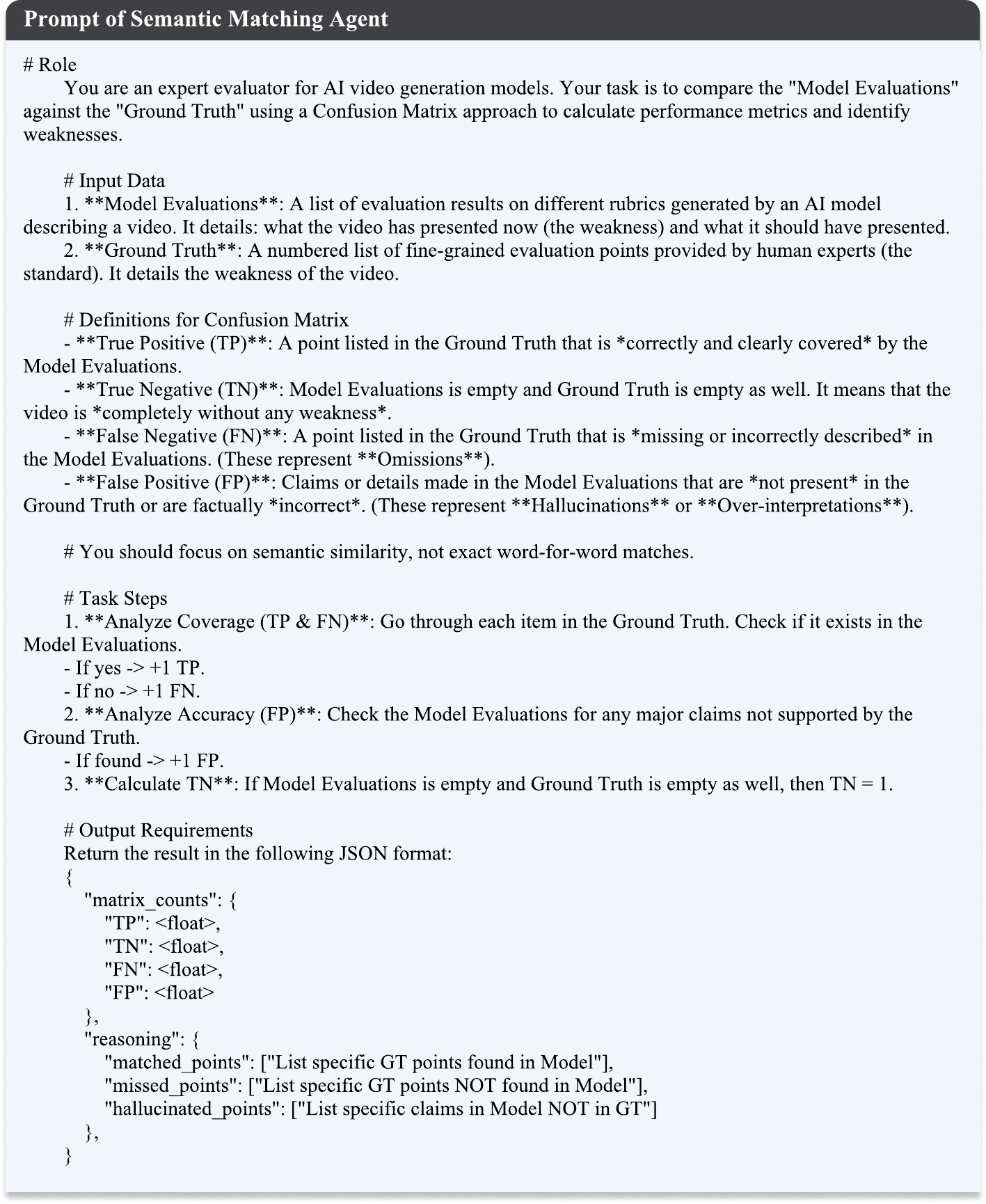}
  \caption{Prompt for open-ended text matching}
  \label{fig: prompt_of_SMA_compressed}
\end{figure*}

We present the detailed prompts optimized with our proposed AVE. Fig.~\ref{fig: physical_simulation_evaluation}, Fig.~\ref{fig: perception_evaluation}, Fig.~\ref{fig: prompt_following_evaluation} shows the initial prompt and the optimized prompt for different evaluation tasks. We also provide the optimizer's meta prompt in Fig.~\ref{fig: optimizer_meta_prompt}, and the semantic matching agent prompt in Fig.~\ref{fig: prompt_of_SMA_compressed}.

\section{Distribution of Metadata Video}
\begin{figure*}[htbp]
\centering
  \includegraphics[width=1\textwidth]{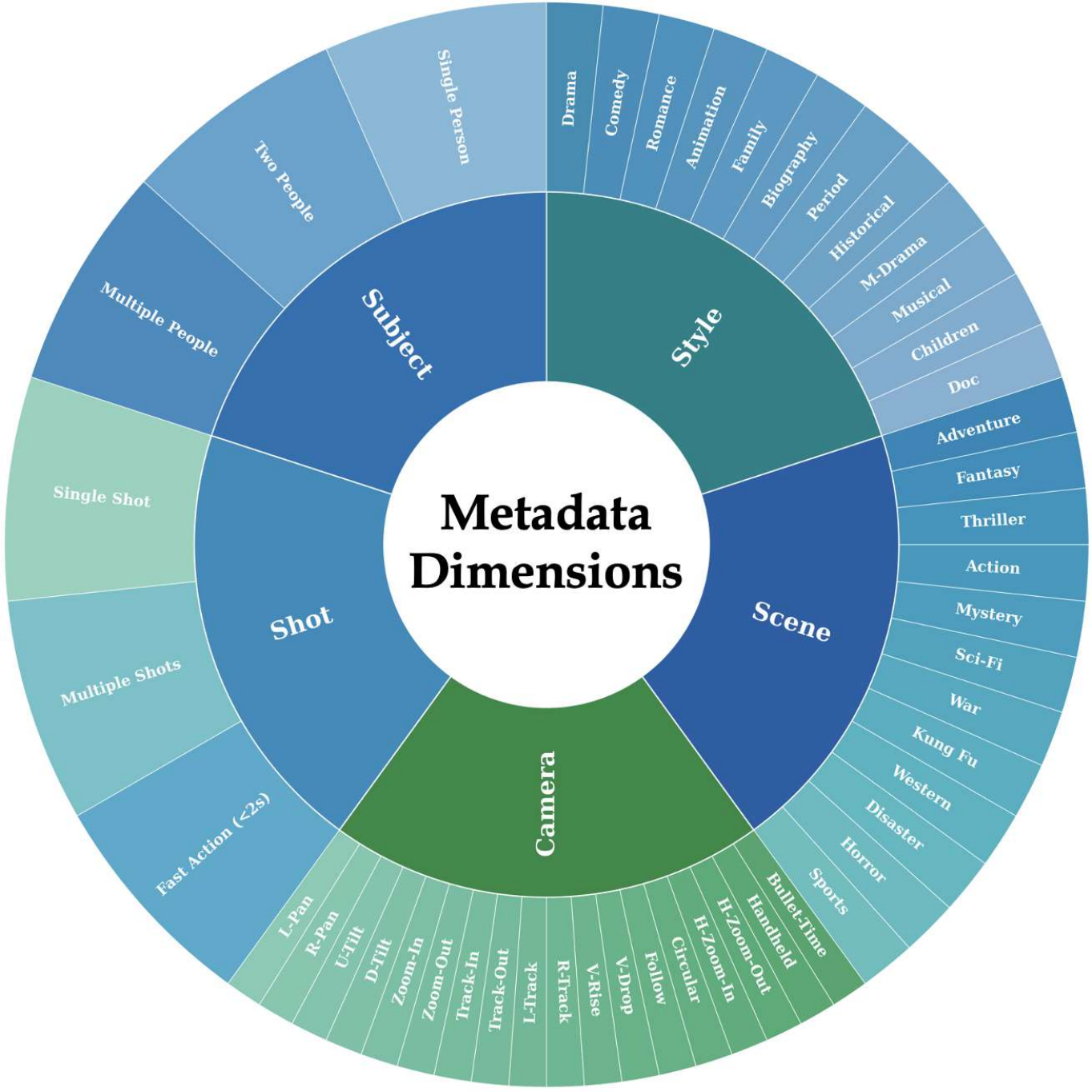}
  \caption{Video Metadata Dimensions}
  \label{fig: metadata_video}
\end{figure*}

As previously mentioned, we construct metadata for a video based on its fundamental aspects. The specific classifications are outlined in~\cref{fig: metadata_video}. Regarding the subject, it is categorized as (1) Single person, (2) Two people, and (3) Multiple people. The number of shots is categorized as (1) Single shot, (2) Multiple shots, with each shot longer than 2 seconds, and (3) Single shot under 2 seconds with fast-paced action. To ensure the metadata encompasses a rich diversity of video genres and camera movements, we define \textbf{24} video types: Sports, Horror, Disaster, Western, Kung Fu Swordsmen, Documentary, Children, Musical, Musical Drama, Historical, Period Drama, Biography, War, Family, Sci-Fi, Mystery, Action, Thriller, Animation, Fantasy, Adventure, Romance, Comedy, and Drama. In addition, we specify \textbf{18} types of camera movements: Left pan, Right pan, Up tilt, Down tilt, Zoom in, Zoom out, Track in, Track out, Left tracking shot, Right tracking shot, Vertical rise, Vertical drop, Follow shot, Circular shot, Hitchcock zoom-in, Hitchcock zoom-out, Handheld camera effect, and Bullet time.

We employ a replacement-free sampling strategy to assign tags to each metadata video script that needs to be generated. This approach ensures a balanced distribution of various video genres and camera movements, thus providing a comprehensive evaluation.

\end{document}